\documentclass[lettersize,journal]{IEEEtran}
\usepackage{amsmath,amsfonts}
\usepackage{algorithmic}
\usepackage{algorithm}
\usepackage{array}
\usepackage{booktabs}
\usepackage[caption=false,font=normalsize,labelfont=sf,textfont=sf]{subfig}
\usepackage{textcomp}
\usepackage{stfloats}
\usepackage{multirow}
\usepackage{url}
\usepackage{verbatim}
\usepackage{graphicx}
\usepackage{amssymb}
\usepackage{booktabs}
\usepackage{amssymb}
\usepackage{bm}
\usepackage{makecell}
\usepackage[table]{xcolor}
\usepackage[colorlinks,linkcolor=blue]{hyperref}
\usepackage{rotating}
\usepackage{ulem}

\begin{document}
\title{Drone Referring Localization: An Efficient Heterogeneous Spatial Feature Interaction Method For UAV Self-Localization}

\author{
Ming Dai,
Enhui Zheng,
Jiahao Chen, 
Lei Qi,
Zhenhua Feng,~\IEEEmembership{Senior~Member,~IEEE} and 
Wankou Yang,~\IEEEmembership{Member,~IEEE}
\IEEEcompsocitemizethanks{\IEEEcompsocthanksitem M. Dai, W. Yang are with the School of Automation, Southeast University, Nanjing 210096, China. E-mails: (mingdai, wkyang)@seu.edu.cn. \IEEEcompsocthanksitem E. Zheng, J. Chen are with the Unmanned System Application Technology Research Institute, China Jiliang University, Hangzhou, 310018, China. Email: (ehzheng, P21010854010)@cjlu.edu.cn. \IEEEcompsocthanksitem Z. Feng is with the School of Artificial Intelligence and Computer Science, Jiangnan University, Wuxi 214122, China. E-mail: fengzhenhua@jiangnan.edu.cn \IEEEcompsocthanksitem L. Qi is with the School of Computer Science, Southeast University, Nanjing 210096, China E-mails: qilei@seu.edu.cn}
}

\maketitle

\begin{abstract}

Image retrieval (IR) has emerged as a promising approach for self-localization in unmanned aerial vehicles (UAVs). However, IR-based methods face several challenges: 1) Pre- and post-processing incur significant computational and storage overhead; 2) The lack of interaction between dual-source features impairs precise spatial perception. In this paper, we propose an efficient heterogeneous spatial feature interaction method, termed Drone Referring Localization (DRL), which aims to localize UAV-view images within satellite imagery. Unlike conventional methods that treat different data sources in isolation, followed by cosine similarity computations, DRL facilitates the learnable interaction of heterogeneous features. To implement the proposed DRL, we design two transformer-based frameworks, Post-Fusion and Mix-Fusion, enabling end-to-end training and inference. Furthermore, we introduce random scale cropping and weight balance loss techniques to augment paired data and optimize the balance between positive and negative sample weights. Additionally, we construct a new dataset, UL14, and establish a benchmark tailored to the DRL framework. Compared to traditional IR methods, DRL achieves superior localization accuracy (MA@20 +9.4\%) while significantly reducing computational time (1/7) and storage overhead (1/3). The dataset and code will be made publicly available.
The dataset and code are available at \url{https://github.com/Dmmm1997/DRL} .

\end{abstract}


\begin{figure}[h]
 	\centering
 	\includegraphics[width=0.95\linewidth]{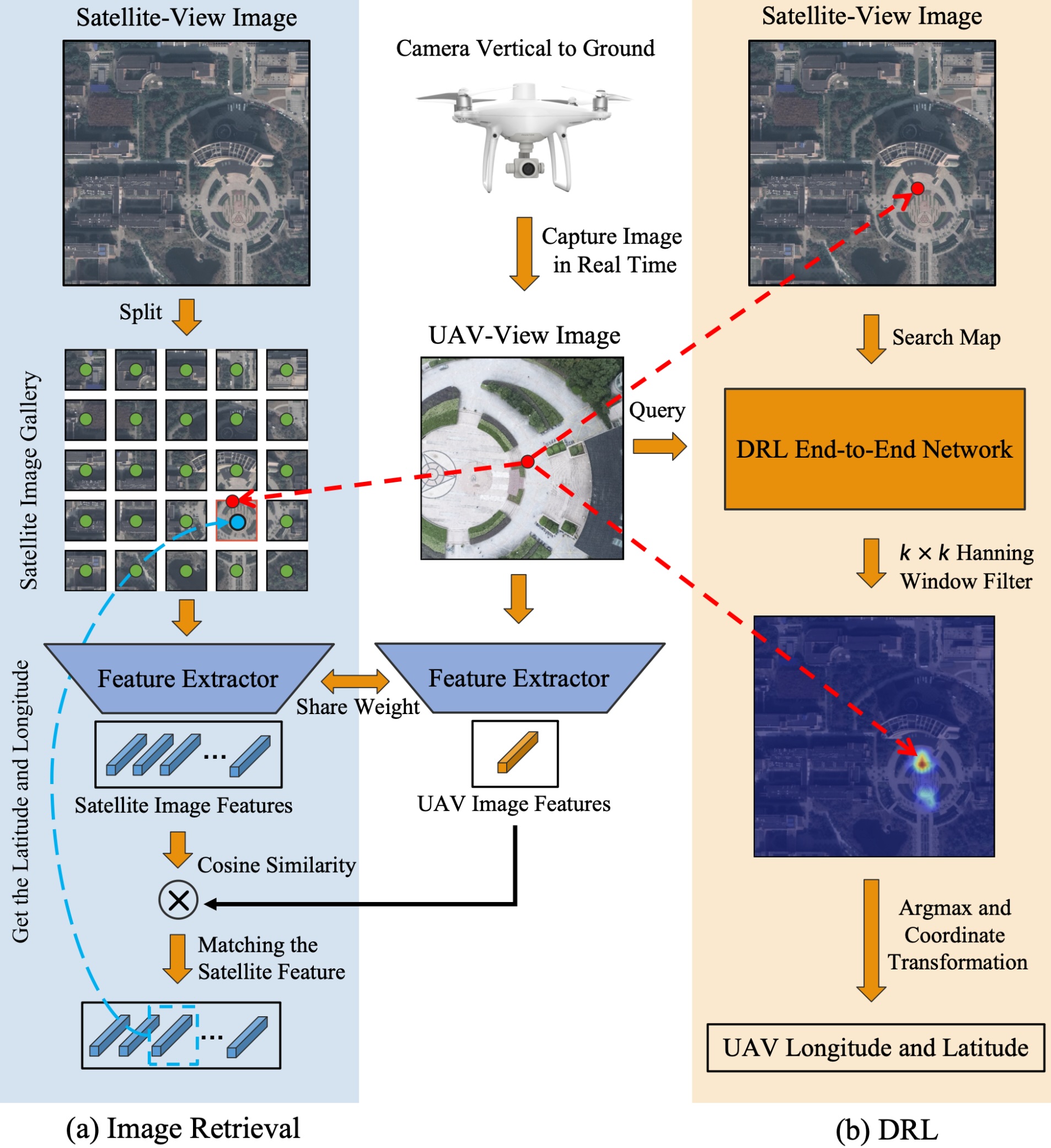}
 	\caption{A comparison between (a) image retrieval and (b) the proposed DRL framework for UAV self-localization. In IR-based methods, the features of different satellite images are spatially isolated and do not interact with UAV image features. Additionally, this approach requires complex pre- and post-processing operations. The green dot represents the sampling position in the gallery, so the IR-based method inevitably introduces inherent errors caused by sampling. In contrast, the proposed DRL method adopts an end-to-end heterogeneous spatial feature fusion architecture, which simplifies the entire localization process and overcomes the inherent errors.
  }
 	\label{figure_retrival_vs_FPI}
    \vspace{-5pt}
\end{figure}

\section{Introduction}\label{sec1}
\IEEEPARstart{I}{n} recent years, UAVs have seen extensive application in vision-related tasks across various domains~\cite{bib1, tits1, tits2}, such as facility inspection, agricultural operations, ground reconnaissance, and civilian aerial photography. UAVs typically rely on global positioning system (GPS) signals for self-localization. However, the stability of these signals varies across different environments, and UAVs often operate in scenarios where GPS signals are weak or completely unavailable. In such situations, UAVs may experience operational failures or even complete loss of control. This paper explores an alternative self-localization approach for UAVs that relies solely on visual information, providing a viable solution in GPS-denied environments.

For conventional positioning tasks, feature point matching methods such as scale-invariant feature transform (SIFT)~\cite{SIFT} and speeded-up robust features (SURF)~\cite{SURF} are commonly used. These methods rely on hand-crafted feature descriptors that exhibit advantageous properties, including scale, rotation, and invariance. They have been widely and successfully applied in tasks like simultaneous localization and mapping (SLAM)~\cite{mur2015orb} and image stitching~\cite{brown2007automatic}. However, in scenarios where image quality varies significantly across domains, corner-point matching methods can only extract a limited number of distinctive features, leading to a substantial decline in performance. To address these limitations, deep learning has emerged as a powerful alternative, capable of capturing richer feature relationships and delivering superior performance in challenging tasks.

The UAV self-localization task presents several challenges, including drastic domain differences, uncertain scale, inconsistent viewing angles, and misaligned spatial information due to time offsets. 
Thus, some deep learning methods have been developed to improve content understanding and feature integration. 
One such method is DenseUAV~\cite{bib16}, which adopted an image retrieval scheme and used the UAV-view image as a query to retrieve the most similar satellite-view image from a satellite gallery. 
Subsequently, GPS information is obtained to determine the UAV's position. 
At present, fine-grained positioning methods have emerged for matching ground panoramas with satellite images~\cite{xia2022visual,xia2023convolutional,lentsch2023slicematch,fervers2023uncertainty}, with VIGOR-type~\cite{VIGOR_2021} methods being prominent examples. These methods improve positioning accuracy by adding offset prediction to image retrieval. This paper focuses on the UAV self-localization task, aiming to achieve end-to-end UAV positioning through an efficient method distinct from image retrieval. In this context, the process of searching for a rough position can be bypassed. Due to the spatial continuity of object movement, once the initial position is determined, a corresponding rough satellite image can be cropped. If the UAV's location can be identified within this rough satellite image, its trajectory can be continuously tracked as it moves by iteratively applying this process.


Although IR-based methods significantly improve positioning accuracy, they have limitations in two key aspects. 
(1) These methods do not integrate features from different sources, limiting their ability to evaluate image similarity based solely on feature distance from respective sources. As a result, IR-based methods are unable to learn spatial relative relationships and offer an isolated solution for UAV self-localization. Furthermore, because the gallery in IR-based methods is sampled at fixed spatial intervals, data-level errors are inevitable. As illustrated by the red dotted line in Fig.~\ref{figure_retrival_vs_FPI}, the green points represent the positions of the cropped satellite images. If the sampling interval is large, meaning the green points are sparse, positioning errors will increase. However, denser sampling leads to higher computational and storage costs.
(2) IR-based methods cannot achieve end-to-end processing during application, as they involve multiple steps with high time and space complexity, as shown in Fig.~\ref{figure_retrival_vs_FPI} (a).
First, in the pre-processing step, large-scale satellite images need to be cropped and a feature gallery needs to be constructed through model inference. 
The calculation and storage consumption of this part depend on the sampling interval and the size of the real space area to be retrieved.
Then, in the post-processing part, the UAV image features acquired in actual scenes need to be compared against the satellite gallery by calculating the cosine distance between features.
The computational burden here is directly related to the size of the gallery. Although optimization methods like KNN~\cite{knn} can reduce computational load, they also introduce a certain loss of accuracy, which is suboptimal for tasks requiring high precision.


Given the limitations discussed above, this paper proposes an end-to-end heterogeneous spatial feature interaction method for UAV self-localization, termed Drone Referring Localization (DRL). For clarity, we refer to the UAV image as the \textit{query} and the satellite image as the \textit{search map}. The process is illustrated in Fig.~\ref{figure_retrival_vs_FPI} (b). DRL simultaneously inputs both drone and satellite images into the model and outputs a heatmap whose spatial distribution aligns with the satellite image. Each pixel in the heatmap represents the probability of the drone being at the corresponding location. 
Unlike IR-based methods, DRL allows for full interaction between heterogeneous source features and avoids errors related to sampling intervals, thereby enhancing positioning accuracy. Additionally, DRL streamlines the entire localization process by eliminating the need for complex pre- and post-processing operations, reducing computational and storage overhead, and improving inference speed.

The primary focus of this paper is to investigate an end-to-end heterogeneous feature interaction method for UAV self-localization. Additionally, we conduct experiments from both data and model perspectives to validate the proposed approach. To support the training requirements of the DRL architecture, we reconstruct the pairwise attribute dataset, UL14, derived from the DenseUAV classification dataset, with a more challenging test set configuration. Furthermore, we propose two model architectures, Post-Fusion and Mix-Fusion, which integrate heterogeneous features at the end and intermediate stages of the backbone network, respectively.
To comprehensively evaluate performance, we design two additional metrics: Meter-level Accuracy (MA), which intuitively measures spatial distance, and Relative Distance Score (RDS), which assesses positioning performance at the model level. Furthermore, we apply data augmentation techniques, including random scaling and random offset, to diversify the training sample pairs. Additionally, a negative sample weight factor is introduced within the balanced loss framework, significantly improving optimization during training.

The main contributions can be listed as follows: 
\begin{enumerate}
    \item We propose an efficient heterogeneous feature interaction method for UAV self-localization task called Drone Referring Localization (DRL), which deeply interacts with heterogeneous features in a learnable manner and circumvents complex pre- and post-processing steps.
    \item We present two model architectures, Post-Fusion and Mix-Fusion, and conclude that the Mix-Fusion is more effective than the Post-Fusion. Moreover, we introduce a data augmentation technique involving random scale and offset to enrich the diversity of training data. Additionally, we improve balance loss and analyze the impact of the number and weight of positive and negative samples.
    \item We constructed a new benchmark, which includes a new dataset UL14 constructed from paired samples, and evaluation indicators Meter-level Accuracy (MA) and Relative Distance Score (RDS). 
    \item The proposed DRL has higher positioning accuracy (MA@20 +9.4\%) while significantly reducing time (1/7) and storage (1/3) overhead than the IR-based solution.
\end{enumerate}
The remainder of this paper is structured as follows. 
First, we introduce the related work in Section~\ref{sec2}. 
Then, the model architecture of DRL is designed in Section~\ref{sec4}, and the proposed dataset and evaluation metrics are presented in Section~\ref{sec3}. 
Next, we analyze the experiments from data- and model-level in Section~\ref{sec5} and Section~\ref{sec6}. 
And the visualization and limitations are drawn in Section~\ref{sec7} and Section~\ref{sec8}. Last, we conclude the context of this paper in Section~\ref{sec9}

\section{Related Work}\label{sec2}
\subsection{Geo-Localization Dataset}\label{sec2.1}
\subsubsection{Ground-to-Aerial Matching}\label{sec2.1.1}
Geo-localization originally addressed the task of matching ground and aerial imagery. 
Some seminal investigations conducted by~\cite{bib11, bib12, bib13} laid the foundation for leveraging publicly available resources to establish image pairs encompassing ground and aerial views. 
Subsequently, the CVUSA~\cite{bib14} project devised image pairings derived from ground-based panoramic images and satellite imagery, whereas CVACT~\cite{bib15} further enhanced CVUSA by incorporating spatial elements, such as orientation maps. 
Recently, VIGOR~\cite{VIGOR_2021} redefined the problem by adopting a more realistic assumption: that the query image can be of arbitrary nature within the specified area of interest.
\subsubsection{Drone-to-Satellite Matching}\label{sec2.1.2}
Univerisity-1652~\cite{bib1} introduced drone-view into cross-view geo-localization and proposed two drone-based subtasks: \textit{drone-view target localization} and \textit{drone navigation}, and regarded them as image retrieval tasks. 
SUE-200~\cite{SUE-200} improved the model's adaptability to drone flight altitudes by collecting drone images at 4 different altitudes. 
Inspired by Univeristy-1652, DenseUAV~\cite{bib16} achieved high precision localization of UAVs by dense sampling, which is the first time to solve the vision-based UAV self-localization problem by the scheme of image retrieval.

\subsection{Deeply-Learned Geo-Localizatioin}
\label{section 2.2}

Due to the potential applications, Cross-View Geo-Localization (CVGL) has gotten increased attention in recent years. 
Some pioneering methods~\cite{bib34,bib35,bib36,bib37}, concentrated on extracting hand-crafted features. 
Inspired by the powerful feature extraction capability of deep Convolutional Neural Networks (CNNs), most of the later works are carried out on the basis of deep learning~\cite{bib38,bib39,bib45,bib47,OFT}. 
Since drone-view was introduced for the CVGL task, several methods have been proposed specifically for addressing the challenges in drone-view~\cite{zhu2023modern, deuser2023orientation}. 
DSM~\cite{bib48} took into account a limited field of view and employed a dynamic similarity matching module to align the orientation of cross-view images. 
PLCD~\cite{bib49} took advantage of drone-view information as a bridge between ground-view and satellite-view domains. 
LPN~\cite{lpn} proposed the square-ring partition strategy to allow the network to pay attention to more fine-grained information at the edge and achieved a huge improvement. 
FSRA~\cite{bib28} introduced a simple and efficient transformer-based structure to enhance the ability of the model to understand contextual information as well as to understand the distribution of instances.
MCCG~\cite{mccg} introduced a ConvNeXt-based method to captures rich discriminative information by cross-dimension interaction and acquires multiple feature
representations.



\subsection{Transformer in Vision}
\label{section 2.3}
In recent years, Transformer~\cite{bib17} has gradually become the mainstream basic model architecture in the fields of natural language processing (NLP)~\cite{bib18,albert} and computer vision (CV)~\cite{bib19,bib20}. 
Some CVGL methods have also been further improved with the help of Transformer. 
Currently, the backbone based on the vision Transformer can be mainly divided into two categories. 
One is the unstructured Transformer structure, the most typical models such as ViT~\cite{bib19}, and DeiT~\cite{bib20}. 
The other is the hierarchical Transformer structure, which typically includes PvT~\cite{PVT_2021}, CvT~\cite{CvT}, and Swin-Transformer~\cite{bib21}, which are widely used in fine-grained tasks such as object detection~\cite{detr, zhu2020deformable} and image segmentation~\cite{strudel2021segmenter,SETR}.

Considering the requirements of this task for spatial context understanding and fine-grained mining capabilities. 
Through the experience of previous work~\cite{bib16} and the verification of our experiments, Transformer can perform better. 
More details will be analyzed in Section~\ref{sec6.4}.


\begin{figure*}
	\centering
	\includegraphics[width=1.0\linewidth]{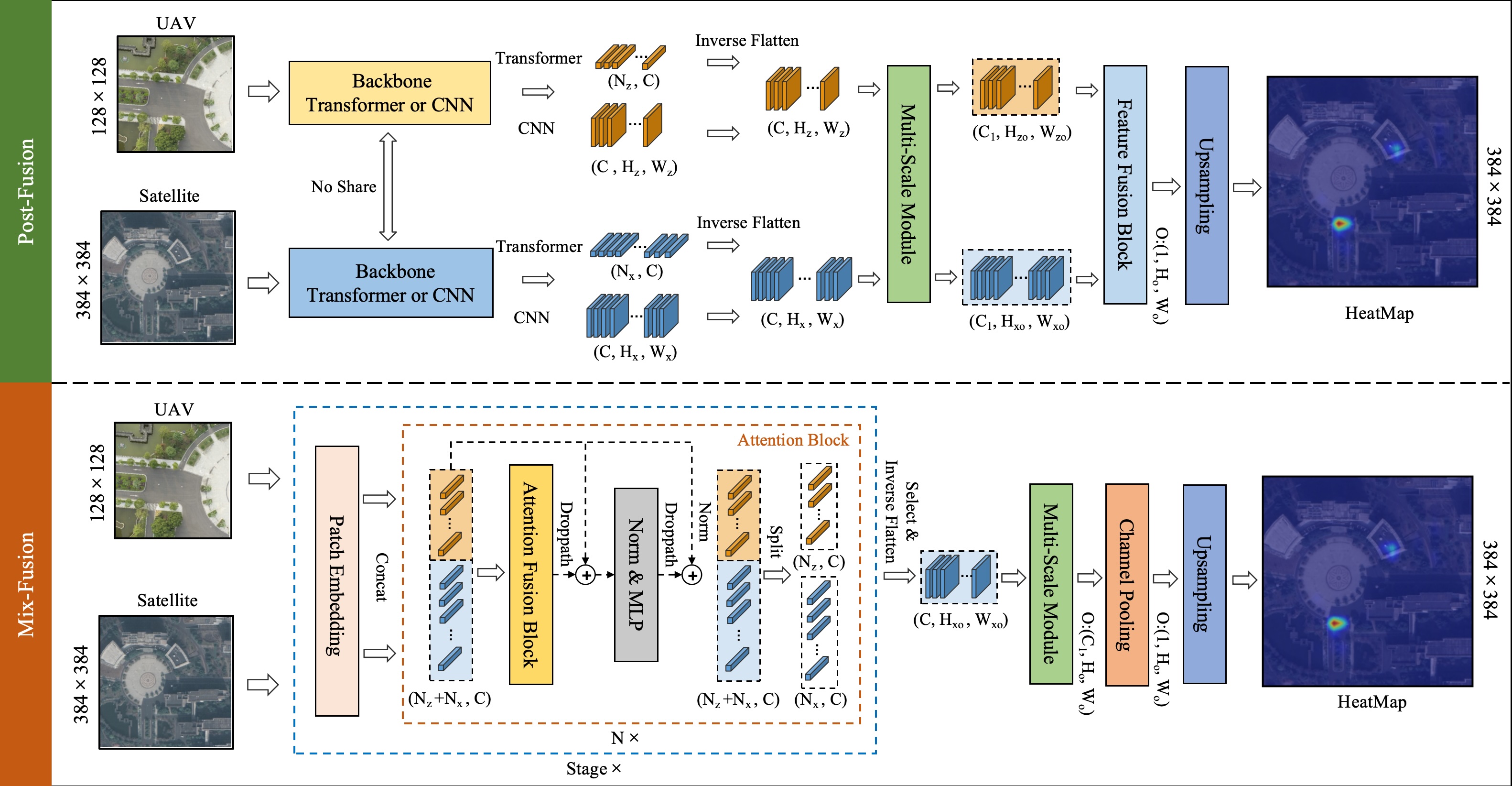}
        \vspace{-10pt}
	\caption{Post-Fusion is a dual-stream architecture that interacts with UAV- and satellite-view features through a Feature Fusion Block. Mix-Fusion is a single-stream network that interacts with feature information in the feature extraction part.}
	\label{figure_network}
        \vspace{-10pt}
\end{figure*}


\section{Methodology}\label{sec4}
\subsection{Localization Schemes}\label{sec4.1}
The proposed framework, Drone Referring Localization (DRL), introduces a novel approach to UAV self-localization. 
To the best of our knowledge, it is the first architecture to employ an end-to-end solution for this task. 
To clearly convey the innovative aspects of DRL, we provide a brief overview of the IR-based scheme in Section~\ref{sec4.1.1}, followed by an in-depth exploration of the unique features of DRL in Section~\ref{sec4.1.2}.

\subsubsection{Image Retrieval}\label{sec4.1.1}

This approach estimates image similarity by comparing features across domains, then indirectly derives GPS information for localization. 
The detailed process is illustrated in Fig.~\ref{figure_retrival_vs_FPI}(a). 
However, IR-based methods have several inherent limitations: 
(1) \textit{Isolated heterogeneous features}: UAV and satellite images lack interaction. Similarity is only assessed by the distance between isolated final features, without integrating information from different sources. 
(2) \textit{Inherent positioning error}: As shown by the green sampling points in Fig.~\ref{figure_retrival_vs_FPI}(a), the sampling interval introduces inevitable error, even for the optimal retrieval result, which deviates from the ground truth. 
(3) \textit{Complicated pre-processing}: The interval for satellite image cropping directly affects positioning accuracy and speed; finer cropping improves accuracy but reduces inference speed. 
(4) \textit{Intricate post-processing}: Calculating feature similarity incurs substantial computational costs, which increase linearly with the size of the satellite gallery. 
(5) \textit{Inflexible model iterability}: When the model is updated, the feature gallery must be regenerated. 
These limitations hinder IR-based methods from achieving real-time inference performance.

\subsubsection{Drone Referring Localization (DRL)}\label{sec4.1.2}
The feature similarity calculation process in image retrieval can be analogized to convolution operations, where feature correlation is computed using a sliding window. 
Inspired by this approach and drawing on techniques from the SOT field, which typically handles inputs from the same domain, we propose a novel localization framework named Drone Referring Localization (DRL) to address the UAV self-localization task. 
The DRL architecture not only facilitates effective interaction between heterogeneous features, resulting in significantly improved positioning accuracy, but also distinguishes itself from IR-based methods by eliminating the need for complex pre- and post-processing steps. 
This allows UAV self-localization to be accomplished through end-to-end inference. 
The steps for achieving localization with the DRL framework are illustrated in Fig.~\ref{figure_retrival_vs_FPI}(b). 
DRL performs feature interaction, fuses information from the two sources, and ultimately outputs a heatmap representing the spatial probability distribution. 
Following simple post-processing steps, such as applying a Hanning window filter and coordinate transformation, the UAV's position can be determined. 
DRL not only simplifies the UAV self-localization process but also significantly improves inference speed and reduces storage resource consumption, which will be further analyzed in Section~\ref{sec5.2}.


\begin{figure*}
	\centering
	\includegraphics[width=1.0\linewidth]{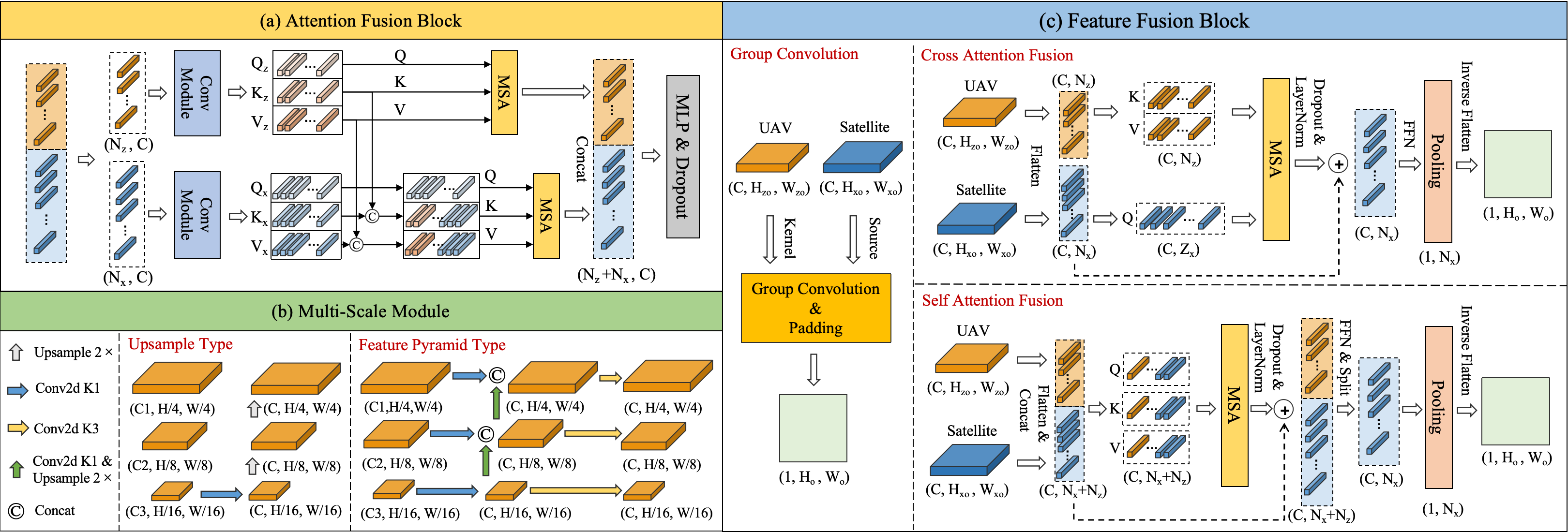}
        \vspace{-10pt}
	\caption{This diagram illustrates some detailed modules in the DRL architecture (Fig. \ref{figure_network}). (a) is the Attention Fusion Block part of the Mix-Fusion structure. The Multi-Scale Module represented by (b) can be applied to both architectures, and two types of structures are used here to implement it. (c) is the Feature Fusion Block part of the Post-Fusion structure, which contains 3 different types to fuse heterogeneous features.}
	\label{figure_blocks}
        \vspace{-10pt}
\end{figure*}

\subsection{Model Structure of DRL}\label{sec4.2}
This section offers an elaborate explanation of two model structures designed for DRL architecture. The first structure employs a dual-stream network, named \textit{Post-Fusion} (refer to Section~\ref{sec4.2.1}). The second structure utilizes a single-stream network, named \textit{Mix-Fusion} (refer to Section~\ref{sec4.2.2}).

\subsubsection{Post-Fusion}\label{sec4.2.1}

As illustrated in Fig.~\ref{figure_network}, the Post-Fusion entails a dual-stream structure, where images from two sources undergo feature extraction and subsequent fusion using feature fusion blocks (FFB). The feature extraction phase commonly employs backbones such as CNNs or Transformers, which will be discussed in Section~\ref{sec6.4}. 

This paper introduces three feature fusion approaches, namely group convolution (GC), cross-attention fusion (CAF), and self-attention fusion (SAF). Their structures are shown in Fig.~\ref{figure_blocks}(c). The experimental analysis of FFB will be discussed in Session~\ref{sec6.1}.

\textit{Group Convolution (GC)}: The features of the UAV-view are treated as the kernels, while the features of the satellite-view serve as the sources. Interaction between the features of the two domains is achieved through group convolution~\cite{gc}.

\textit{Cross-Attention Fusion (CAF)}: The spatial dimensions of UAV and satellite features are flattened to $\textit{N}_\textit{z}=\textit{H}_{\textit{zo}}\times\textit{W}_{\textit{zo}}$ and $\textit{N}_\textit{x}=\textit{H}_{\textit{xo}}\times\textit{W}_{\textit{xo}}$ respectively.
The features from UAV-view $F_z\in\mathbb{R}^{(\textit{C}, \textit{N}_\textit{z})}$ serve as key and value, while the features from satellite-view $F_x\in\mathbb{R}^{(\textit{C}, \textit{N}_\textit{x})}$ serve as query. The multi-head self-attention (MSA) mechanism is employed to fuse the features, producing the corresponding features $F_o\in\mathbb{R}^{(\textit{C}, \textit{N}_\textit{x})}$ of the same size as query. Then, dropout and layernorm operations are applied to $F_o$ and added to $F_x$ to resemble a residual structure. The FFN module is used for feature integration, involving linear, activation, normalization, and dropout operations. Last, the C-dimension is compressed through average pooling and restored to the width and height shape (1, $\textit{H}_\textit{o}$, $\textit{W}_\textit{o}$) via an inverse flatten operation.

\textit{Self-Attention Fusion (SAF)}: SAF constructs query, key, and value by concatenating the features from the UAV- and satellite-view. Specifically, the input dimension of the multi-head attention mechanism is ($C$, $N_z$+${N}_{x}$). Last, the part originally belonging to the satellite feature is divided from the spliced features through the split operation.


\subsubsection{Mix-Fusion}\label{sec4.2.2}

Mix-Fusion is essentially a single-stream network structure that allows image features from two domains to fuse within the transformer backbone, without additional parameters. Therefore, the pre-trained weights can still be loaded. The structure is shown in the lower half of Fig.~\ref{figure_network}. First, images from two sources are simultaneously fed to the model, and compressed by patch embedding to 1/4 of the original size. Next, flatten the spatial dimension as $F_z\in\mathbb{R}^{(\textit{N}_\textit{z}, \textit{C})}$ and $F_x\in\mathbb{R}^{(\textit{N}_\textit{x}, \textit{C})}$. Then, concatenate $F_x$ and $F_z$ to obtain $F_m\in\mathbb{R}^{(\textit{N}_\textit{x}+\textit{N}_\textit{z}, \textit{C})}$ and use the attention fusion block (AFB) for feature fusion. Here, a residual structure is added to protect the original features and prevent the attention mechanism from degrading the results. Then, normalize the features through layernorm and enrich feature diversity through MLP. The original input, after normalization, is added to the output to obtain the new fused features $F_m$. Last, complete one stage by splitting $F_m$ into $F_z$ and $F_x$. Excluding the patch embedding operation, the remaining part constitutes a complete attention block process. 

Next, we introduce the structure of the AFB, as depicted in Fig.~\ref{figure_blocks}(a). 
First, the input features are divided into $F_z$ and $F_x$, which are then mapped to query, key, and value using a convolutional (conv) module. 
Specifically, the conv module performs inverse flatten, convolution, and flatten operations. 
The attention mechanism in the AFB consists of two parts: 
(1) a self-attention mechanism for UAV features, and 
(2) a cross-attention mechanism where the satellite features provide the query, while the concatenated features from both domains serve as the key and value. 
Finally, the outputs representing the UAV and satellite features are concatenated, followed by MLP and dropout operations.

After the feature extraction process, the satellite-view features are restored to their original dimensions $(\textit{C}, \textit{H}_\textit{xo}, \textit{W}_\textit{xo})$ using an inverse flatten operation. 
A multi-scale module (MSM) is optionally applied to capture both shallow and deep features. 
Additionally, channel pooling is utilized to compress the channel dimension, and upsampling is employed to restore the spatial scale.

\subsection{Feature Scales}\label{sec4.3}

\subsubsection{Multi-Scale Modules (MSM)}\label{sec4.3.1}
We adopt two resolution expansion strategies to evaluate the impact of multi-scale processing on UAV self-localization tasks. 
First, the upsampling strategy, as depicted in Fig.~\ref{figure_blocks}(b), restores the scale by applying bilinear interpolation to the final layer feature, which has dimensions $(\textit{C3}, \textit{H/16}, \textit{W/16})$. 
Second, the feature pyramid approach~\cite{fpn}, also shown in Fig.~\ref{figure_blocks}(b), leverages the original structure of the feature pyramid, merging features from different layers through a combination of convolution, upsampling, and concatenation operations. 
The experimental results of these strategies will be discussed in Section~\ref{sec6.3.1}.

\subsubsection{Output Resolution}\label{sec4.3.2}
For the UAV self-localization task, the resolution of the output feature map significantly impacts the localization accuracy. A higher resolution leads to a smaller localization error. In fact, experimental results demonstrate that adding an upsampling module at the end of the model effectively improves positioning performance. Further analysis is provided in Section~\ref{sec6.3.2}.

\subsection{Random Scale Crop (RSC)}\label{sec4.4}

\begin{figure}
	\centering
	\includegraphics[width=0.9\linewidth]{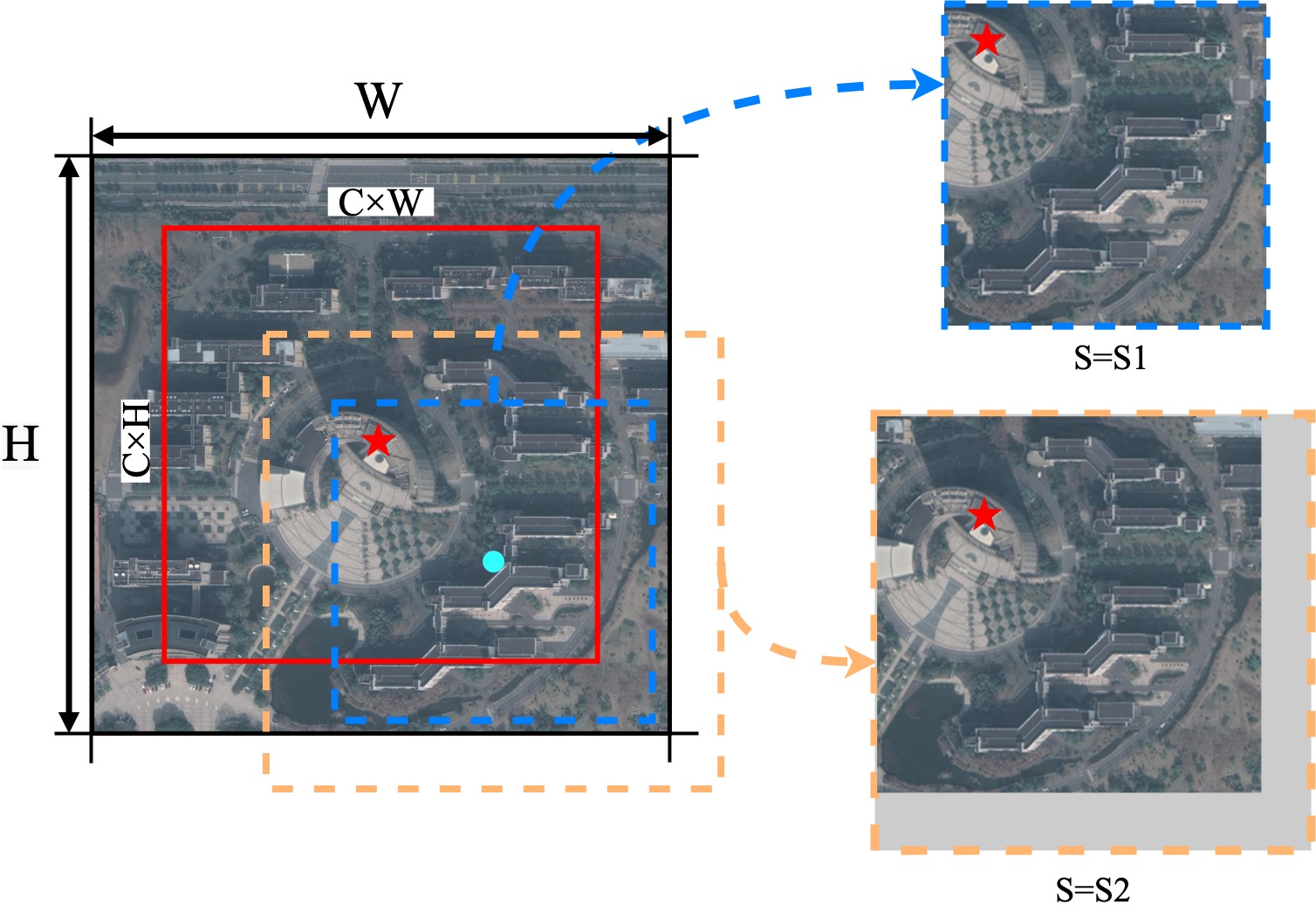}
	\caption{The process of the RSC augmentation method. The red pentagon is the position of the UAV. The light blue dot is the randomly generated center position of the image. The orange and blue dotted boxes correspond to 2 cropped images of different scales.}
	\label{figure_data_aug}
        \vspace{-10pt}
\end{figure}

There are two primary data construction schemes for training DRL: pre-building fixed UAV-satellite pairs and generating pairs randomly through data augmentation. We adopt the latter approach for greater flexibility and introduce a random scale and offset augmentation method, termed Random Scale Crop (RSC). 

The RSC method generates sample pairs with both random target distributions and random scales. Figure~\ref{figure_data_aug} illustrates the implementation of RSC, which uses two hyperparameters to dynamically crop satellite-view images during training: the centroid coverage range \textit{C} (default 0.85) and the scale range \textit{S} (default 512$\rightarrow$1000). 
The hyperparameter \textit{C} controls the distribution interval of the query within the search map. A smaller \textit{C} results in queries being concentrated closer to the center of the search map. The hyperparameter \textit{S} determines the spatial scale of the search map, with a larger \textit{S} increasing the spatial distance covered by each pixel after resizing. 
The RSC method aims to enhance the model's robustness to scale and offset variations. In simpler terms, \textit{C} defines the size of the red area where queries can be distributed, while different values of \textit{S} (e.g., S1 and S2) are used to generate search maps at varying scales.

\begin{figure}
	\centering
	\includegraphics[width=1.0\linewidth]{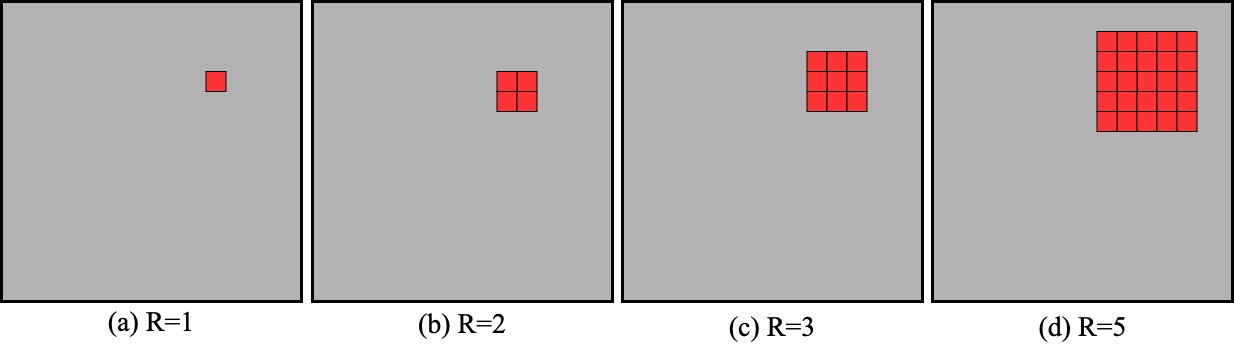}
	\caption{Schematic diagram of selecting positive samples according to \textit{R}.}
	\label{fig_centerR}
        \vspace{-10pt}
\end{figure}
\begin{algorithm}[!h] 
	\renewcommand{\algorithmicrequire}{\textbf{Input:}}
	\renewcommand{\algorithmicensure}{\textbf{Output:}}
	\caption{Weighted Balance Loss.} 
	\label{alg1} 
	\small
	\begin{algorithmic}[1]
		\REQUIRE $\textit{map}\in{\mathbb{R}^{\textit{H}\times\textit{W}}}$, $\textit{label}\in{\mathbb{R}^{\textit{H}\times\textit{W}}}$, 
		$\textit{w}_\textit{neg}\in{\mathbb{R}}$, 
		$\textit{R}\in{\mathbb{N}^+}$
		\ENSURE loss   \\
		\# Generate the 0,1 matrix as shown in Fig. \ref{fig_centerR}
		\STATE $(\textit{map},\textit{label},R)\Rightarrow t$ \\
		\# Num of the positive and negative samples
		\STATE $\textit{w} \gets \textit{t}$ 
		\STATE $\textit{N}_\textit{pos} = \textit{R}^2$ \\
		\STATE $\textit{N}_\textit{neg} = \textit{H}\times\textit{W}-\textit{R}^2$  \\
		\# Weight of the positive and negative samples
		\STATE $\textit{W}_\textit{pos} = 1/\textit{N}_\textit{pos}$  \\
		\STATE $\textit{W}_\textit{neg} = (1/\textit{N}_\textit{neg})\times{\textit{w}_\textit{neg}}$ \\
		\# Weight normalization
		\STATE $\textit{w} \gets \textit{w}/\sum{\textit{w}_\textit{i}}$ \\
		\# Map normalization
		\STATE $\textit{p} \gets \textit{sigmoid}(\textit{map})$ \\
		\# Weighted balance loss
		\STATE $\textit{loss} = -\sum({{\textit{log}(\textit{p}_\textit{ij})\textit{t}_{ij} \textit{w}_\textit{ij}}+{\textit{log}(1-\textit{p}_\textit{ij})(1-\textit{t}_\textit{ij})\textit{w}_\textit{ij}}})$\\
		\RETURN \textit{loss}
	\end{algorithmic}
\end{algorithm}

\subsection{Weighted Balance Loss (WBL)}\label{sec4.5}

The number of positive and negative samples, as well as their associated weights, are crucial variables in model optimization. An imbalance in these samples can lead to the model converging towards an extreme mode that minimizes the loss function, which may result in misguiding the model or slowing down convergence. balance loss~\cite{bib8} is an effective method for balancing positive and negative samples, with its core principle being to maintain a 1:1 weight ratio between them. However, balance loss does not sufficiently address the issue of sample imbalance, and a strict 1:1 balance is not optimal for localization tasks. 
To address these issues, we propose an improved method called weighted balance loss (WBL), which is detailed in Alg.~\ref{alg1}. WBL classifies samples within a specified range \textit{R} around the ground truth as positive samples and introduces a weight factor, $\textit{w}_\textit{neg}$, to adjust the influence of negative samples.

Specifically, the parameter \textit{R} is chosen based on the number of positive samples. As illustrated in Fig.~\ref{fig_centerR}, \textit{R=1} implies that only the position closest to the ground truth is considered positive, with all other positions classified as negative. When \textit{R=2}, the four neighboring points around the ground truth are also considered positive samples. As \textit{R} increases, the number of positive samples increases accordingly, but the weight assigned to each positive sample decreases proportionally.
 
 
In terms of sample weights, firstly, the number of positive and negative samples should be counted. A weight matrix whose size is the same as the heatmap needs to be prepared to store the weight of each sample. Since the number of negative samples is much larger than that of positive, the weight of each positive is much larger than that of the negative after balancing. Therefore, $\textit{w}_\textit{neg}$ is introduced to dynamically adjust the weights for negative samples. Last, the weight matrix is multiplied by the cross-entropy loss value for all samples.

 \section{Dataset and Evaluation}\label{sec3}

\begin{table*}[h]
\centering
\caption{
    Geo-localization dataset summary. The "Training" column indicates the number of samples in each location. The UL14 training set is constructed using 1-to-1 paired data.
}
\tiny
\label{tab_dataset_list}
\renewcommand\arraystretch{1.2}
\resizebox{1.0\hsize}{!}{
	\begin{tabular}{ccccccc}
		\specialrule{0.75pt}{0pt}{1pt}
		{Datasets}& 
		{\bf{UL14 (ours)}}&
		{DenseUAV~\cite{bib16}} &
		{SUES-200~\cite{SUE-200}} &
		{University-1652~\cite{bib1}}& 
		{VIGOR~\cite{VIGOR_2021}}&
		{CVUSA~\cite{bib14}} 
		\\
		\specialrule{0.5pt}{1pt}{1pt}
		{Training} & 
		{6.8k$\times$2}&
		{10$\times$225.6$\times$9}& 
		{120$\times$51}&
		{701$\times$71.64}& 
		{91k+53k}&
		{35.5k$\times$2} 
		\\
		{Platform} & 
		{Drone, Satellite}&
		{Drone, Satellite}&
		{Drone, Satellite}&
		{Drone, Ground, Satellite}&
		{Ground, Satellite}&
		{Ground, Satellite} 
		\\
		{Imgs./Platform}&
		{1 + 1}& 
		{3 + 6} & 
		{50 + 1} & 
		{54 + 16.64 + 1} & 
		{/}&
		{1 + 1}
		\\
		{Target}&
		{UAV}&
		{UAV}&
		{Diverse}&
		{Building}&
		{User}&
		{User}
		\\
		{Method}& 
		{DRL}&
		{IR}& 
		{IR}& 
		{IR}& 
		{IR + Reg}&
		{IR} 
		\\
		{Evaluation}&
		{RDS \& MA}&
		{R@K \& SDM}& 
		{R@K \& AP \& RB \& PF}& 
		{R@K \& AP}& 
		{MA}&
		{R@K}
		\\
		\specialrule{0.75pt}{1pt}{0pt}
	\end{tabular}
        \vspace{-10pt}
}
\end{table*}

\begin{table}[h]
	\centering
	\caption{Statistical table for the UL14 dataset, where \#Universities refers to the number of universities included.}
	\tiny
	\label{tab_UL14}
	\renewcommand\arraystretch{1.1}
	\resizebox{0.9\hsize}{!}{
		\begin{tabular}{cccc}
			\specialrule{0.75pt}{0pt}{0.5pt}
			\multirow{2}{*}{\#Subset} & \multicolumn{2}{c}{\#Imgs\ \ \ } & \multirow{2}{*}{\#Universities} \\ \cline{2-3}
			& UAV-view & Satellite-view & \\
			\specialrule{0.5pt}{0.5pt}{0.5pt}
			Train & 6768 & 6768 & 10 \\
			Test & 2331 & 27972 & 4 \\
			\specialrule{0.75pt}{0.5pt}{0pt}
		\end{tabular}
	}
        \vspace{-10pt}
\end{table}

\subsection{Dataset Description}
\label{sec3.1}

\begin{figure*}
	\centering
	\includegraphics[width=1.0\linewidth]{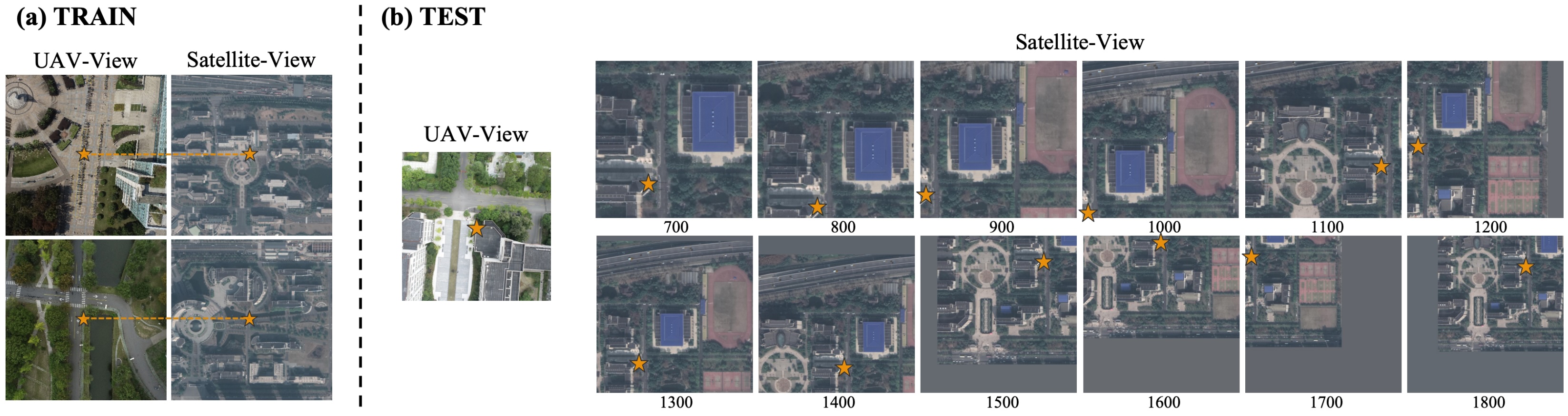}
	\caption{Visualization of some samples in the train and test set. The numbers in the test set represent the original pixel scale (0.294 meters/pixel).}
	\label{figure_dataset_visual}
        \vspace{-10pt}
\end{figure*}

Table~\ref{tab_dataset_list} lists the amount of training data, sampling platforms, data distribution, localization targets, and evaluation metrics for some typical geo-localization datasets. Explicitly, the UL14 dataset possesses the following characteristics. (1) Image from two perspectives: UAV and satellite views. (2) Paired training data: the UAV images and satellite images in the training data are constructed in pairs, and the data augmentation method will be used in the actual training process to expand the diversity of paired data. (3) Multiple flight altitudes: UAV images are sampled in three flight altitudes (80, 90, and 100 meters). (4) Multi-scale test data: The test satellite data is constructed of 12 different scales to enhance the difficulty of positioning.

Next, we will provide a detailed introduction to the construction of the train and test set, and the quantity statistics are presented in Table~\ref{tab_UL14}.

\subsubsection{Train Set}\label{sec3.1.1}
The training set comprises images collected by 10 universities. To enhance the utility of the training set, the resolution of UAV-view images is uniformly set to (512, 512), while the satellite images are set to (1280, 1280). The decision to retain larger satellite images is made to allow for greater flexibility in subsequent data augmentation, which is be discussed in Section~\ref{sec4.4}. Last, some paired images of the train set are shown in Fig.~\ref{figure_dataset_visual}(a).

\subsubsection{Test Set}\label{sec3.1.2}
The test set comprises images collected by 4 universities, with no overlap with the train set.  The spatial scale of the satellite images is defined by setting the pixels in the range of 700-1800 (0.294 meter/pixel) at intervals of 100 pixels. This implies that each UAV image in the test set generated 12 satellite images of varying scales. Also, the position of the UAV in the satellite imagery is randomly distributed, either in the center or at the edge. For a clear view, one set of images in the test set is shown in Fig.~\ref{figure_dataset_visual}(b).


\subsection{Evaluation Indicators}\label{sec3.2}
Recall@K and AP\cite{recall, ap} are commonly used evaluation metrics, which only consider whether the sample is positive or negative. This is a kind of discrete metric, which defeats the real purpose of the localization task. Unlike the indirect positioning of the image retrieval scheme, DRL needs a more intuitive way to evaluate the accuracy of localization. Therefore, we propose two evaluation metrics, Meter-level Accuracy (MA) and Relative Distance Score (RDS), to evaluate the localization accuracy from the space- and model-level respectively.

\label{section 3.4}
\subsubsection{Meter-level Accuracy (MA)}\label{sec3.2.1}
One intuitive way to measure localization accuracy is to analyze it from the level of spatial distance, that is, to measure the Meter-level Accuracy (MA). The expression of the MA evaluation index proposed in this paper is as follows:
\begin{equation}
	\text{MA@K}=\frac{\sum_{i=1}^{N}{1_{\text{SD}<\text{Km}}}}{N}
	\label{eq_MA}
\end{equation}
\begin{equation}
	1_{\text{SD}<\text{Km}} =  \begin{cases}1 & \text{SD} < \text{Km}\\0 & \text{SD}  \geq  \text{Km}\end{cases}
	\label{eq_1SD}
\end{equation}
where SD refers to the real spatial distance in meters. K is an adjustable parameter.  MA@K refers to the number of samples with positioning errors within Km as a proportion of the total samples. In conclusion, MA@K can be represented as the accuracy of positioning error less than Km.  The expression of SD is as follows:
\begin{equation}
	\text{SD}=\sqrt{(\Delta{x})^2+({\Delta{y}})^2}
	\label{eq_SD}
\end{equation}
$\Delta{x}$ denotes the meter-level error between the model prediction and the ground-truth in the longitude direction, and $\Delta{y}$ denotes the meter-level error in the latitude direction.

\begin{figure}
	\centering
	\includegraphics[width=0.5\linewidth]{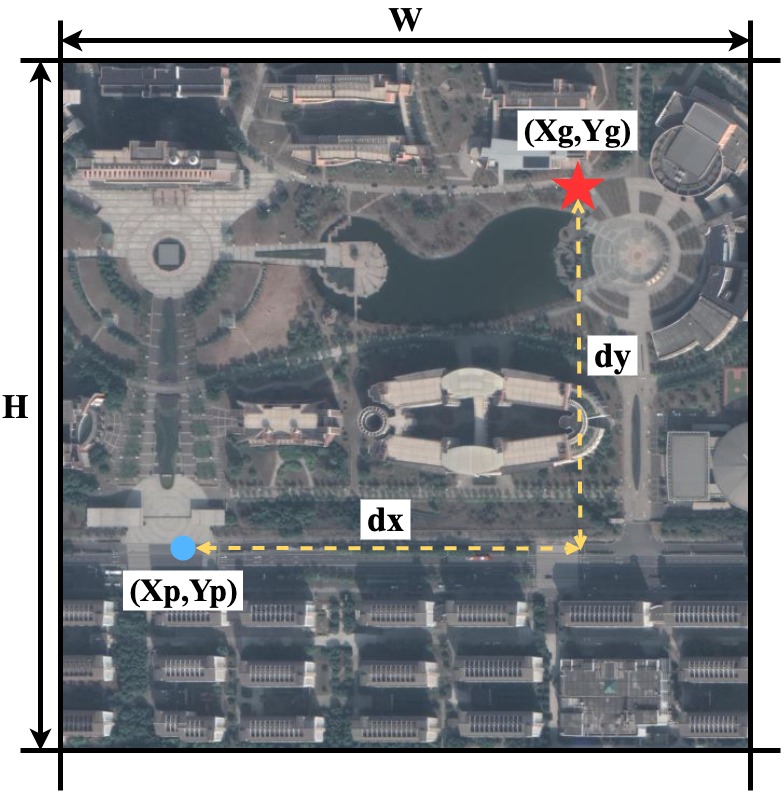}
	\caption{The pentagram ($X_g$,$Y_g$) indicating ground-truth and the blue circle ($X_p$,$Y_p$) indicating the position predicted by the model.}
	\label{figure_RDS_blanceloss}
 \vspace{-10pt}
\end{figure}

\subsubsection{Relative Distance Score (RDS)}\label{sec3.2.2}
Although MA can express the accuracy of positioning intuitively, it also has some shortcomings. 1) The setting of K has different effects on search maps of different scales. For large-scale search maps, after resizing, the spatial distance represented by a unit pixel will also become larger. Therefore, there will be a small deviation in the search map, but a large deviation in the spatial distance. 2) MA often needs to be expressed by a series of K, which is not convenient for directly expressing the performance of the model.

Based on the above discussion, we propose the Relative Distance Score (RDS) to evaluate from the model level. As shown in Fig.~\ref{figure_RDS_blanceloss}, \((X_p,Y_p)\) represents the predicted coordinates and \((X_g,Y_g)\) represents the ground-truth coordinates. The expression for Relative Distance (RD) is as follows:
\begin{equation}
	\text{RD}=\sqrt{\frac{(\frac{dx}{w})^2+(\frac{dy}{h})^2}{2}}
	\label{eq1}
\end{equation}
where $dx=\left|X_p-X_g\right|$ and $dy=\left|Y_p-Y_g\right|$. To make the RD perform in the same direction as the accuracy and distributed between 0 and 1, we convert the distance into a score, and the expression of the converted RDS is as follows:
\begin{equation}
	\text{RDS}=e^{-k\times\sqrt{\frac{(\frac{dx}{w})^2+(\frac{dy}{h})^2}{2}}}
	\label{eq5}
\end{equation} 
where $k$ is the scaling factor, which is set to 10 by default. 

Why propose an RDS? First of all, RDS measures the performance of the model by the relative distance at the pixel level in the image, which can resist the scale transformation of the search map. The distance in RDS is no longer the real space distance, but the pixel distance in the image, which is beneficial to evaluate the performance from the model level. Second, the RDS is an overall score, which is distributed between 0 and 1, and is positively correlated with the positioning accuracy, that is, the higher the RDS, the higher the positioning accuracy. Finally, RDS uses an exponential method to expand the relative distance. When the distance increases, the RDS will return to 0 in an exponential form, which is in line with the expectation that the large distance deviation will be regarded as a wrong positioning.

\section{Experiment}\label{sec5}

\subsection{Implementation Details}\label{sec5.1}
In the experiments, all initial parameters of backbones are pre-trained on ImageNet~\cite{ImageNet}. 
The AdamW~\cite{adamw} optimizer is adopted with a weight decay of 5e-4. Moreover, we incorporate a cosine annealing learning rate decay schedule, and the minimum learning rate is 1/100 of the initial learning rate.  
During the training phase, the UAV-view images are set at a resolution of 128$\times$128 pixels, while the satellite-view images are formatted at 384$\times$384 pixels in default.
For the inference phase, the hanning window size is the same as the training setting of \textit{R} in WBL. 

\subsection{DRL v.s. Image Retrieval}\label{sec5.2}

\subsubsection{Localization Performance}\label{sec5.2.1}


To make the comparison as fair as possible, we uniformly use the real space meter-level accuracy (MA@K) to compare these two schemes. Since the test set of DenseUAV is constructed at intervals of 20m and the gallery contains images that perfectly match each query. We have additionally constructed a gallery of satellite images with a fixed sampling interval and not perfectly center-corresponding, which is consistent with the setting of the UL14. We have used some previous SOTA models in image retrieval for comparative experiments. The experimental results are shown in Table~\ref{tab_FPI_retrival}. The Mix-Fusion structure of the proposed DRL architecture has an absolute improvement of 9.4 points over the IR-based FSRA model on the MA@20 indicator. In high-precision positioning indicators such as MA@3, there is also an absolute improvement of 5.2 points.


\subsubsection{Efficiency}\label{sec5.2.2}
The detailed time consumptions of DRL and image retrieval are counted in Table \ref{tab_time_storage_consumption}, which contains 600 samples. In order to ensure the fairness of the experimental data as much as possible, all operations are performed without using multi-process technology. The main steps are as follows:
\begin{itemize}
\item Building Satellite Gallery (BSG): Intensive split of satellite images to construct a gallery database just like the process in Fig.~\ref{figure_retrival_vs_FPI}(a). The IR-based method takes about 4.2s, while DRL only requires a simple crop operation, which only takes microseconds.
\item Gallery Forward Propagation (GFP): Extract features for all images in the gallery. The consumption of the IR-based scheme is determined by both the sampling interval and the gallery size, while DRL does not require this step.
\item Inference And Localization (IAL): Real-time inference and localization. For IR-based methods, Part of the overhead is UAV image inference, which accounts for a relatively small proportion. The main overhead comes from the calculation of feature similarity between UAV and satellite gallery. For DRL, only model inference is needed.
\end{itemize}
the DRL architecture has achieved significant advantages in inference through the end-to-end architecture.
Even without considering the time-consuming preprocessing of BSG and GFP, the DRL method is still more than 7 times faster than the IR-based method.


\begin{table}[!t]
	\tiny
	\centering
	\renewcommand\arraystretch{1.0}
	\caption{The positioning accuracy of the DRL method based on the Mix-Fusion architecture is significantly improved compared to the IR-based method. This is mainly attributed to the architectural advantages of DRL, which deeply interacts with heterogeneous features and overcomes the inherent error problem.}
	\label{tab_FPI_retrival}
	\resizebox{1.0\hsize}{!}{
		\begin{tabular}{c|c|c|ccc}
			\specialrule{0.75pt}{0pt}{1pt}
			Type & Method & Backbone &MA@3 &MA@10&MA@20   \\
			\specialrule{0.5pt}{1pt}{1pt}   
			\multirow{8}{*}{\begin{sideways}Image Retrieval\end{sideways}}&Uni-1652~\cite{bib1}&RN50&0.005&0.047&0.145\\
                &Triplet Loss~\cite{tripletloss}&RN50&0.006&0.046&0.140\\
			&LCM~\cite{LCM}&RN50&0.014&0.112&0.250\\
                &LPN~\cite{lpn}&RN50&0.015&0.158&0.273\\
                &RKNet~\cite{RKNet}&RN50&0.021&0.177&0.317\\
			&Uni-1652~\cite{bib1}&ViT-S&0.069&0.541&0.723\\
			&GeM~\cite{gem}&ViT-S&0.076&0.526&0.707\\
                &MCCG~\cite{mccg}&ConvNext-T&0.078&0.574&\underline{\textbf{0.752}}\\
			&FSRA~\cite{bib28}&ViT-S&\underline{\textbf{0.079}}&\underline{\textbf{0.580}}&0.743\\
                &FSRA~\cite{bib28}&CvT13&0.029&0.265&0.423\\
			\specialrule{0.5pt}{1pt}{1pt}   
			\multirow{2}{*}{\begin{sideways}DRL\end{sideways}}&Post-Fusion&CvT13&0.084&0.436&0.643\\
			&Mix-Fusion&CvT13&\underline{\textbf{0.131}}&\underline{\textbf{0.625}}&\underline{\textbf{0.837}}\\
			\specialrule{0.75pt}{1pt}{0pt}
		\end{tabular}
}
\end{table}

\begin{table}[!t]
	\centering
	\renewcommand\arraystretch{1.2}
 	\caption{Time and storage consumption comparison of FSRA and DRL. Without considering preprocessing, DRL is still 7 times faster than IR-based methods and consumes 3 times less storage.}
	\label{tab_time_storage_consumption}
	\resizebox{1.0\hsize}{!}{
 		\begin{tabular}{c|cccc|cccc}
 			\specialrule{0.75pt}{0pt}{0.5pt}
 			\multirow{2}{*}{Method} & \multicolumn{4}{c|}{Time Consumption(s)} & \multicolumn{4}{c}{Storage Consumption(MB)} \\
    &BSG& GFP & IAL& Total & SI & GI & GF & Total \\
 			\specialrule{0.5pt}{0.5pt}{0.5pt}
 			FSRA(ViT-S) & \sout{4.2} & \sout{6.5} & 43.2 &43.2 & \underline{\textbf{4.8}} & \sout{44.6} &10.8 &15.6\\
                \specialrule{0.5pt}{0.5pt}{0.5pt}
 			Post-Fusion & \underline{\textbf{\textasciitilde0}} & \underline{\textbf{0}} & 6.2& \textasciitilde6.2 & \underline{\textbf{4.8}} & \underline{\textbf{0}} & \underline{\textbf{0}} & \underline{\textbf{4.8}} \\
 			Mix-Fusion & \underline{\textbf{\textasciitilde0}} & \underline{\textbf{0}} & \underline{\textbf{6.0}}& \underline{\textbf{\textasciitilde6.0}} & \underline{\textbf{4.8}} & \underline{\textbf{0}} & \underline{\textbf{0}} & \underline{\textbf{4.8}} \\
 			
 			\specialrule{0.75pt}{0.5pt}{0pt}
 		\end{tabular}
   \vspace{-10pt}
 	}
 \end{table}


\subsubsection{Storage Consumption}\label{sec5.2.3}

Next, the DRL and IR-based schemes will be compared from the perspective of storage. The results are shown in Table \ref{tab_time_storage_consumption}. The main steps is as follows: 
\begin{itemize}
\item Satellite Images (SI): Pre-downloaded satellite maps.
\item Gallery Images (GI): The IR-based method requires the additional generation of an image gallery (split from satellite images). The architecture of DRL only requires a satellite image to be loaded into memory.
\item Gallery Features (GF): The IR-based methods need to construct gallery features through model inference for gallery images, while DRL does not require this step.
\end{itemize}
For the IR-based method, the image generated by GI is an intermediate product. To ensure fairness, after removing this part of the consumption, a total of 15.6MB of memory is occupied. Due to its end-to-end characteristics, DRL also greatly reduces storage consumption, requiring only 1/3 of the IR-based method.


\subsection{Impact of Satellite Spatial Scale}
\label{sec5.3}

\begin{figure*}
	\centering
	\includegraphics[width=0.95\linewidth]{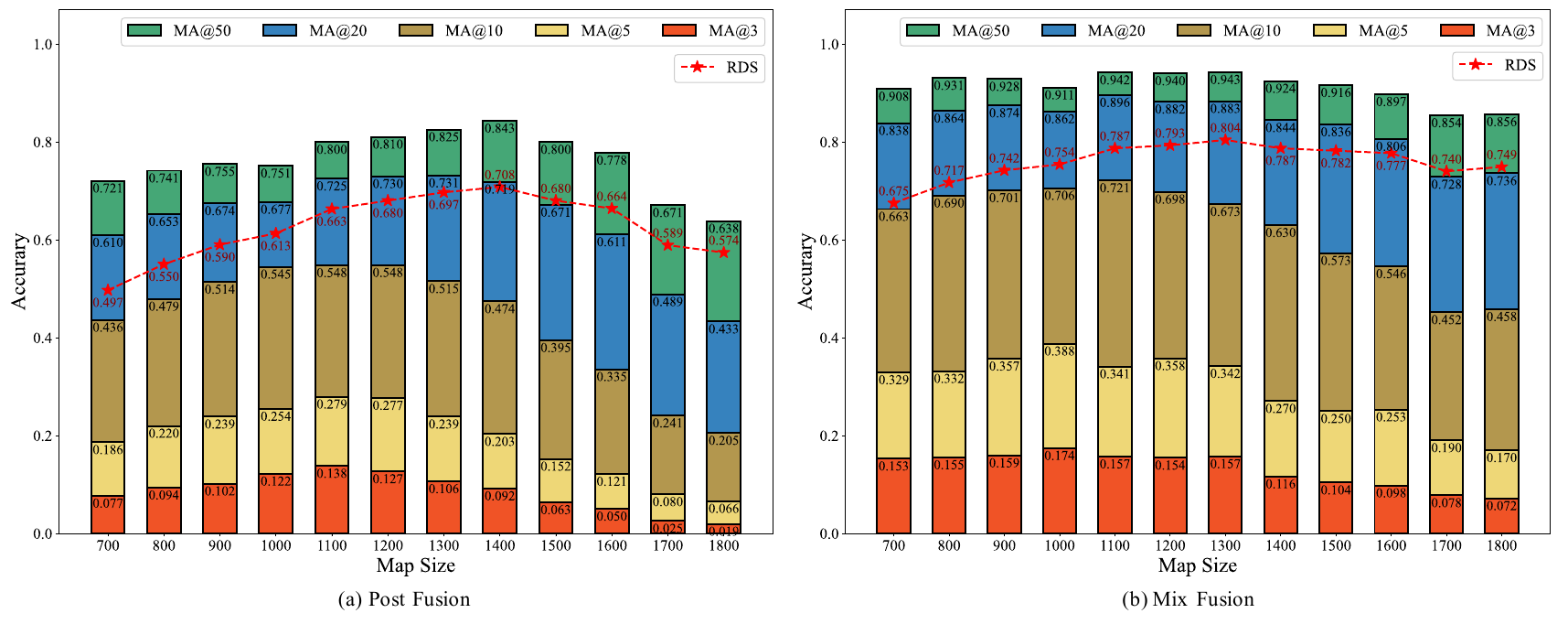}
        \vspace{-10pt}
	\caption{The impact of search mapsize on the positioning of two DRL structures, including RDS and MA@K indicators.}
	\label{figure_spatial_scale}
\end{figure*}


The spatial scale of the search map directly determines the fine-grained information of the input satellite image. Given that the scale of the model input is typically fixed, the model's resistance to scale shifts is important. To investigate the influence of different scales of search maps, we conduct a series of comparison experiments with satellite images ranging from 700 to 1800 pixels (0.294 meters/pixel).

On the one hand, the results of the Post-Fusion structure are presented in Fig. \ref{figure_spatial_scale}(a). We can observe a marked decline in the MA@K index as the scale of the satellite images increases, particularly evident in higher precision indicators like MA@3, MA@5, and MA@10. This trend can be easily rationalized. Given that the input size for all models is consistent, larger satellite image scales would mean that each pixel in the input image represents a larger actual spatial distance. For instance, when the mapsize is 700, the corresponding actual spatial distance is 205.9m. If the output feature map size from the model is 384, then the actual spatial distance between each pixel in the feature map is 0.536m. However, when the mapsize is 1800, it becomes 1.206m.
Although larger mapsize yield poorer localization performance, it is still necessary to use them. The reason lies in the fact that large mapsize can simplify the actual process of UAV self-localization. For instance, if one large image can resolve the localization of all positions, it eliminates the need for any post-processing operations on the application side. 
Further observing the change rule of RDS, we can find that the overall curve is relatively smooth, but it is worth noting that it does not perform well at small scales. This is mainly due to the fact that small-scale images cover less spatial information and the loss of edge information is exacerbated.

On the other hand. the results of the Mix-Fusion structure are shown in Fig. \ref{figure_spatial_scale}(b). It is evident that, compared to the Post-Fusion structure, Mix-Fusion exhibits better robustness to scale. Although it also follows the aforementioned trend, but smoother. This is primarily because deeper information interaction and fusion can assist the model in better learning spatial distribution information.


\subsection{Impact of Query Distribution}
\label{sec5.4}

\begin{figure*}
	\centering
	\includegraphics[width=0.9\linewidth]{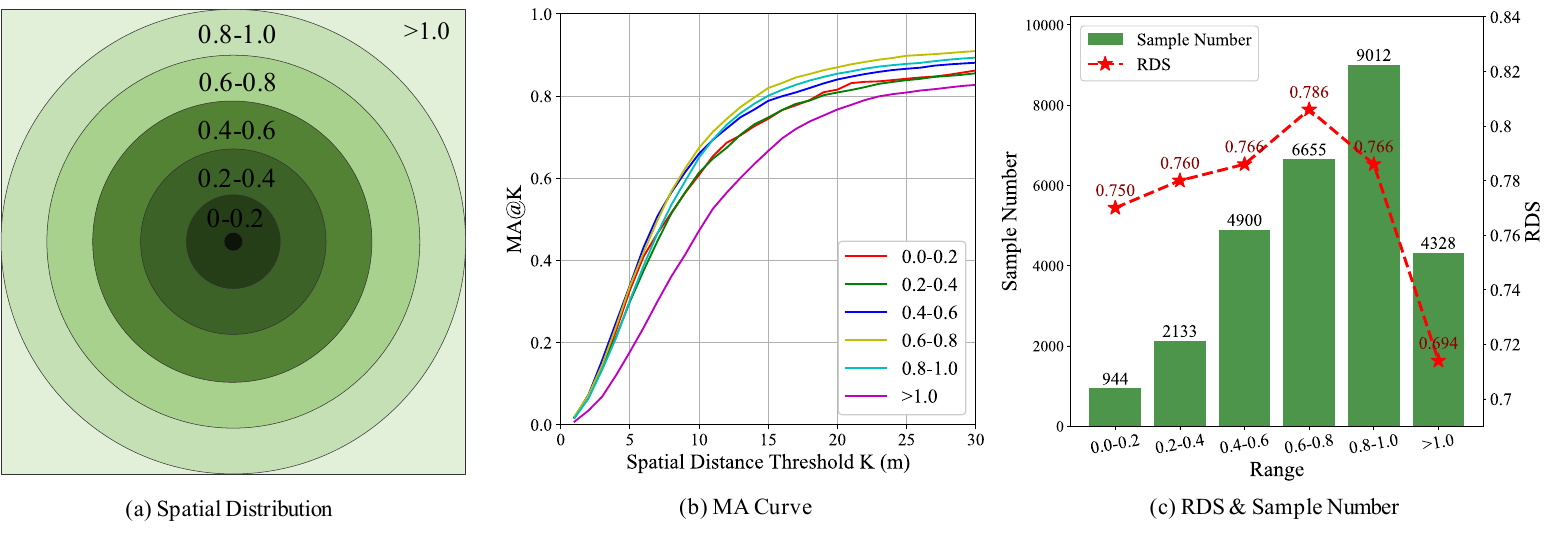}
        \vspace{-10pt}
	\caption{(a) Query distributions. (b) MA curves under different spatial distributions. (c) RDS and sample number under different spatial distributions.}
	\label{figure_query_distribution}
        \vspace{-10pt}
\end{figure*}

During the application, the relative position of the query in the search map is uncertain due to factors such as time delays and flight speed. Therefore, it is important to explore the impact of query distribution. As shown in Fig. \ref{figure_query_distribution}(a), we divide different ranges according to the relative spatial distance between the query and the center point of the search map. The MA curve is shown in Fig. \ref{figure_query_distribution}(b). We can clearly see that there is a significant decrease in localization accuracy when queries are located in the region \textgreater1.0. This is consistent with our expectations, to take an extreme example, when the query is located on 4 corners of the search map, only 1/4 of the information will be available, and the remaining 3/4 of the image information will not find the corresponding information on the search map, which undoubtedly increases the challenge of localization. 
Furthermore, Fig. \ref{figure_query_distribution}(c) illustrates the number of samples, as well as the RDS for each region. The quantity distribution is approximately proportional to the area of the region, which results in fewer samples in the 0-0.2 range and more in the 0.8-1.0 range. From the RDS, we can similarly observe a cliff-like decline in the results for the \textgreater1.0 region. This decline is attributable to the loss of edge information, which greatly affects the discriminative ability of the model. In conclusion, queries located on the edge of the search map significantly reduce localization accuracy.

\subsection{Impact of Flight Altitude}
\label{sec5.5}

\begin{figure}[h]
	\centering
	\includegraphics[width=0.95\linewidth]{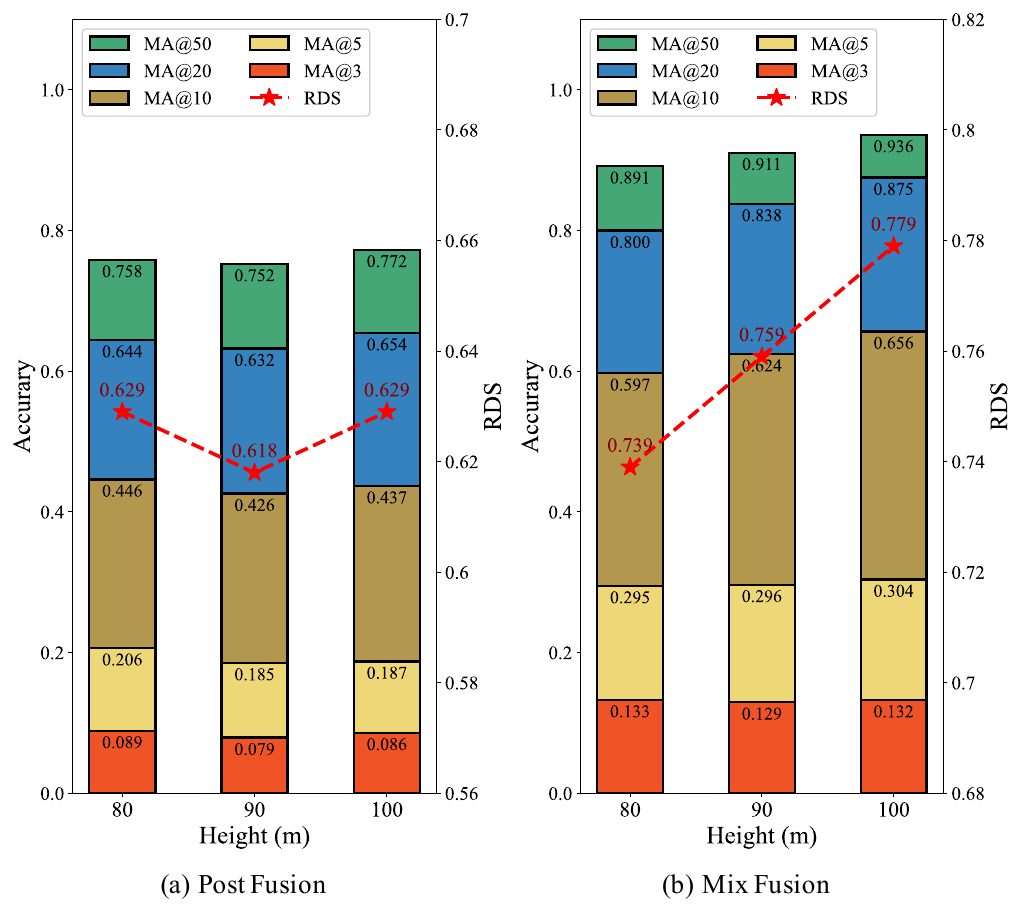}
        \vspace{-10pt}
	\caption{MA@K and RDS of UAV at different flight altitudes.}
	\label{figure_drone_height}
\end{figure}

In the application scenario, the altitude of the UAV flight is an uncertain factor, which directly affects the spatial scale of the query. Taking this into account, UL14 collected UAV images from 80, 90, and 100m altitudes. In the testing phase, data of different heights are divided for separate testing.
The experimental results of the two architecture models are shown in Fig. \ref{figure_drone_height}. It can be seen that both MA@K and RDS show relatively stable performance at these three different heights. However, due to factors such as the height limit of the UAV, higher data have not been collected to fully verify the robustness of the model to flight heights.

\section{Model Analysis}\label{sec6}

In this section, we will conduct experiments and analysis at the model level. Firstly, in Section \ref{sec6.1}, we will systematically analyze the two proposed model architectures. And RDS experiments were introduced in Section \ref{sec6.2}. Then, in Section \ref{sec6.3}, we will introduce two multi-scale approaches and analyze their experimental results. Subsequently, we will specifically analyze the impact of the backbone network and the loss function in Sections \ref{sec6.4} and \ref{sec6.5}. Last, we will analyze some model-related hyperparameters, including input scale, weight sharing, padding, and present the relevant content in Sections \ref{sec6.6}, \ref{sec6.7}, and \ref{sec6.8}, respectively.

\subsection{Model Stucture}\label{sec6.1}

\begin{table*}[h]
	\tiny
	\centering
	\renewcommand\arraystretch{1.0}
	\caption{The impact of fusion methods on the two architectures. For Post-Fusion, three fusion methods were used for ablation experiments corresponding to those described in Section \ref{sec4.2.1}. The experimental content of Mix-Fusion corresponds to the Channel Pooling part in Fig. \ref{figure_network}.}
	\label{tab_structure}
	\resizebox{1.0\hsize}{!}{
		\begin{tabular}{c|c|ccc|c|ccccc}
			\specialrule{0.75pt}{0pt}{0.5pt}
			Structure & Fusion Method & Macs & Params & InferTime & RDS  &  MA@3 & MA@5 &  MA@10 & MA@20 & MA@50\\
			\specialrule{0.5pt}{0.5pt}{0.5pt}
			\multirow{3}{*}{Post-Fusion} & GC & 13.3G & 39.3M & 24.0s & \underline{\textbf{0.576}} & \underline{\textbf{0.043}} & \underline{\textbf{0.111}} & \underline{\textbf{0.329}} & \underline{\textbf{0.581}} & \underline{\textbf{0.727}}\\
			 & SAF & 13.5G & 39.6M &  27.5s & 0.293 & 0.001 & 0.004 & 0.026 & 0.091 & 0.374  \\
			 & CAF & 13.4G & 39.5M & 27.0s & 0.287 & 0.001 & 0.003 & 0.025 & 0.086 & 0.358 \\
			 
			\specialrule{0.5pt}{0.5pt}{0.5pt}
			 
			\multirow{4}{*}{Mix-Fusion} & Mix+AvgPool & \underline{\textbf{13.1G}} & \underline{\textbf{19.6M}} & 22.4s & 0.680 & 0.043 & 0.132 & 0.418 & 0.725 & \textbf{\underline{0.875}}\\
			  & Mix+Linear  & 13.2G & 19.7M & \underline{\textbf{21.3s}} &\textbf{\underline{0.681}}  & 0.045  & 0.137  & \underline{\textbf{0.430}}  & 0.721 &0.873  \\
			  & Mix+GeM &\underline{\textbf{13.1G}} & 19.7M& 21.6s & 0.680 & 0.042 & 0.134 & 0.423 & \textbf{\underline{0.727}} & 0.871 \\
			  & Mix+GC  & 13.2G & \underline{\textbf{19.6M}} & 21.5s & 0.670 & \underline{\textbf{0.056}} & \underline{\textbf{0.146}} & 0.427 & 0.711 & 0.850 \\
			
			\specialrule{0.75pt}{0.5pt}{0pt}
		\end{tabular}
}
\end{table*}


\subsubsection{Post-Fusion}\label{sec6.1.1}

We conducted experiments on the three fusion methods, \textit{GC}, \textit{SAF}, and \textit{CAF}, as shown in Table \ref{tab_structure}. In terms of computation and parameter quantities, the computation of \textit{SAF} is 0.2GMacs higher than that of \textit{GC}, and its parameter quantity is 0.3M larger than that of \textit{GC}, with \textit{CAF} being slightly lower than \textit{SAF}. Regarding inference speed, \textit{SAF} is significantly slower than \textit{GC} and slightly slower than \textit{CAF}. However, in terms of model accuracy, the fusion method of \textit{GC} significantly outperforms the \textit{CAF} and \textit{SAF} methods across all indicators. 
This is a perplexing phenomenon. In the field of SOT, the fusion methods based on attention can interact more fully with features and help alleviate scale inconsistency issues. We attribute this phenomenon to the data level. Given that the input data is from different sources and that there are challenges like rotation uncertainty and spatial information inconsistency due to time, which increases the complexity of learning relationships between features. For attention methods, it means that more training data and iterations are required to learn general knowledge. Conversely, the GC-based method interacts with features using a sliding window, which restricts the spatial scale and makes it easier for the model to learn certain patterns.

\subsubsection{Mix-Fusion}\label{sec6.1.2}

Unlike Post-Fusion, Mix-Fusion can be regarded as a single-stream structure that adds an attention module for feature fusion in blocks of the backbone. As shown in Table~\ref{tab_structure}, compared to the Post-Fusion, Mix-Fusion possesses significant advantages in terms of parameters, speed, and localization accuracy. Specifically, RDS improved from 0.576 to 0.681, a boost of 10.5 points, and MA@20 increased by 14.6\%. This is primarily due to Mix-Fusion enabling deep interaction of context information from the two domains and effectively overcoming the scale inconsistency problem introduced by the GC-based method.
Furthermore, within the framework of Mix-Fusion, we adopted four different methods to compress the number of channels for training: AvgPool, Linear, GeM~\cite{gem}, and GC. As a result, the Linear method achieved the best result on RDS, reaching 0.681, and reached the fastest inference speed of 21.3s/1000samples. It is worth mentioning that the GC method effectively improves the short-distance positioning accuracy, which is the advantage of introducing additional prior inductive bias.

\subsection{RSC Data Augmentation}\label{sec6.2}


In the experiments, three scale augmentation schemes, namely single scale, multiple scale, and random scale are implemented. As shown in Table \ref{tab_augment}, [512, 768] denotes scales of 512 and 768, meanwhile, 512$\rightarrow$768 indicates the random integers down to 512 and up to 768. It can be observed that models trained using single scale underperform in localization tasks. In contrast, both multiple scale and random scale methods result in substantial improvements. This observation aligns with the expectations, as having the model learn universal rules of localization across different scales during training can enhance the model's robustness to scale during testing. Ultimately, our baseline model employs the random scale (512$\rightarrow$1000) augmentation scheme.

\begin{table}[!t]
	\tiny
	\centering
	\renewcommand\arraystretch{1.0}
	\caption{The impact of using different scaling strategies in RSC data enhancement on model performance. Among them, [512, 768] represents the use of two scales, 512 and 768, and 512$\rightarrow$718 represents the use of random scales from 512 to 768.}
	\label{tab_augment}
	\resizebox{1.0\hsize}{!}{
		\begin{tabular}{c|c|c|cc}
			\specialrule{0.75pt}{0pt}{0.5pt}
			
			Scale Type &Scale&RDS&MA@5&MA@20\\
			\specialrule{0.5pt}{0.5pt}{0.5pt}
			
			\multirow{3}{*}{Single Scale} &512 & 0.529 & 0.109 & 0.524 \\
			&768 & 0.533 & 0.081  & 0.489 \\
			&1000 & 0.395 & 0.024  & 0.238 \\
			\specialrule{0.5pt}{0.5pt}{0.5pt}
			
			\multirow{3}{*}{Multiple Scale} &[512, 768] & 0.607 & 0.127 & 0.628 \\
			&[768, 1000] & 0.639 & 0.112  & 0.648 \\
			&[512, 768, 1000] & 0.671 & \underline{\textbf{0.146}}  & 0.703 \\
			\specialrule{0.5pt}{0.5pt}{0.5pt}
			
			\multirow{3}{*}{Random Scale} & 512$\rightarrow$768 & 0.606 & 0.126 & 0.631 \\
			& 768$\rightarrow$1000 & 0.592 & 0.087   & 0.574 \\
			& 512$\rightarrow$1000 & \underline{\textbf{0.680}} & 0.137  & \underline{\textbf{0.721}} \\
		
			\specialrule{0.75pt}{0.5pt}{0pt}
		\end{tabular}
}
\end{table}

\subsection{Feature Scales}\label{sec6.3}

\begin{table*}[h]
	\tiny
	\centering
	\renewcommand\arraystretch{1.0}
	\caption{The impact of different feature scales and feature fusion methods. $\rightarrow$ represents an up-sampling operation.}
	\label{tab_multiscale}
	\resizebox{0.95\hsize}{!}{
		\begin{tabular}{c|c|c|cccccc}
			\specialrule{0.75pt}{0pt}{0.5pt}
			
			Multi-Scale Structure & Output Scale & RDS  &  MA@3 & MA@5 &  MA@10 & MA@20 & MA@50 & MA@100\\
			
			\specialrule{0.5pt}{0.5pt}{0.5pt}
			
			Baseline & (H/16, W/16) &0.680 & 0.045 & 0.137 & 0.430 & 0.721 & 0.873 & 0.939\\
			
			\specialrule{0.5pt}{0.5pt}{0.5pt}
			
			\multirow{4}{*}{Upsample} & (H/8, W/8) & 0.717 & 0.089 & 0.220&0.518&0.773&0.893&0.948\\
			&(H/4, W/4) & \underline{\textbf{0.731}} & 0.100 & 0.244&0.557&\underline{\textbf{0.795}}&\underline{\textbf{0.899}}&\underline{\textbf{0.949}} \\
			&(H/2, W/2) &0.728&0.105 &0.243 &0.547 &0.788 &0.897 &0.948 \\
			&(H/1, W/1)&0.730&\underline{\textbf{0.109}}&\underline{\textbf{0.251}}&\underline{\textbf{0.561}}&0.790&0.893&0.946\\
			
			\specialrule{0.5pt}{0.5pt}{0.5pt}
			
			\multirow{4}{*}{Feature Pyramid} & (H/8, W/8) & 0.733 & 0.114 & 0.270 & 0.584 & 0.796 & 0.844 & 0.940\\
			&(H/4, W/4) & 0.732 & 0.117 & 0.267 & 0.573 & 0.802 & 0.888 & 0.936\\
			&(H/4, W/4) $\rightarrow$ (H/2, W/2) & 0.753 & 0.126 & 0.287 & 0.614 & 0.832 & 0.908 & 0.949\\
			&(H/4, W/4) $\rightarrow$ (H/1, W/1) & \underline{\textbf{0.758}} & \underline{\textbf{0.131}} & \underline{\textbf{0.298}} & \underline{\textbf{0.625}} & \underline{\textbf{0.837}} & \underline{\textbf{0.912}} & \underline{\textbf{0.953}} \\
			
			\specialrule{0.75pt}{0.5pt}{0pt}
		\end{tabular}
	}
\end{table*}


\subsubsection{Multi-Scale Stucture}\label{sec6.3.1}
Experiments for two multi-scale schemes are formulated as shown in Table \ref{tab_multiscale}. Overall, it can be seen that the feature pyramid has achieved noticeable improvements over the brutal upsampling operation. For instance, the RDS of feature pyramid on the (H/1, W/1) scale has increased from 0.730 to 0.758, which is a nearly 2.8 points rise, and an increase of 6.4\% on MA@10. Similarly, there are noticeable improvements on other scales as well. As we know, deep features always contain high-level semantic information, while shallow features carry information such as outline and color. The fusion of shallow features can provide more information for discrimination. However, the significant improvement also demonstrates the necessity of pyramid features in this task.

\subsubsection{Output Resolution}\label{sec6.3.2}
The feature resolution of the final layer based on the CvT13 (\textit{Baseline} in Table \ref{tab_multiscale}) is 1/16 of the original image. In the experimental process, we enlarged the feature scale by a factor of 2 until it matched the original input. However, for the feature pyramid structure, to restore the original scale, we additionally employed bilinear interpolation for up-sampling. As shown in Table \ref{tab_multiscale}, it is evident that with the increase of the output scale, most of the metrics show an upward trend, particularly on the small distance error metrics such as MA@3 and MA@5, which is expected. Due to the enlargement of the output scale, the real spatial distance corresponding to a unit pixel in the output heatmap will become smaller. Therefore, the actual distance deviation caused by wrongly predicting one pixel is multiplicatively reduced. This experiment also verifies a hypothesis that the output scale of the model significantly affects the performance.

\subsection{Backbones}\label{sec6.4}

\begin{table*}[h]
	\tiny
	\centering
	\renewcommand\arraystretch{1.0}
	\caption{The impact of different types of network architectures including CNN and Transformer on the self-localization performance of UAVs.}
	\label{tab_backbone}
	\resizebox{1.0\hsize}{!}{
		
		\begin{tabular}{c|c|c|ccc|c|cccc}
			\specialrule{0.75pt}{0pt}{0.5pt}
			
			Type & Structure &Backbone & Macs & Params & InferTime & RDS  &  MA@3 & MA@5 &  MA@10 & MA@20 \\
			
			\specialrule{0.5pt}{0.5pt}{0.5pt}
			
			\multirow{3}{*}{\begin{sideways}CNN\end{sideways}} & \multirow{3}{*}{Hierarchical} &  ResNet50~\cite{bib10} & 13.7G & 47.2M & 9.9s &  0.208 &  0.001 & 0.004 &  0.016 & 0.059   \\
			&&EfficientNet-B5~\cite{efficientnet} & 7.44G & 54.3M & 33.1s & 0.402  &  0.007 & 0.020 &  0.078 & 0.273   \\
			&&ConvNext-T~\cite{convnext} & 14.6G & 55.7M & \underline{\textbf{8.1s}} & 0.465  &  0.022 & 0.058 &  0.198 & 0.412  \\
			
			\specialrule{0.5pt}{0.5pt}{0.5pt}
			
			\multirow{5}{*}{\begin{sideways}Transformer\end{sideways}}&\multirow{3}{*}{Unstructured}& ViT-S~\cite{bib19} & 13.9G & 43.3M & 9.9s & 0.577  &  0.038 & 0.099 &  0.313 & 0.564  \\
			&&ViT-B~\cite{bib19}& 55.0G & 171.5M & 10.3s & 0.595  &  \underline{\textbf{0.050}} & \textbf{\underline{0.126}} &  \underline{\textbf{0.363}} & 0.597   \\
			&&DeiT-S~\cite{bib20} & 13.9G & 43.3M & 9.7s & \underline{\textbf{0.597}}  &  0.048 & 0.123 &  0.358 & \underline{\textbf{0.604}}  \\
			
			\addlinespace[0.5pt] 
			\cline{2-11}
			\addlinespace[0.5pt]
			
			&\multirow{2}{*}{Hierarchical}&PvT-S~\cite{PVT_2021} & \underline{\textbf{12.0G}}& 47.3M & 18.3s & 0.524 &  0.020 & 0.054&  0.169& 0.470  \\
			& &CvT13~\cite{CvT} & 13.3G & \underline{\textbf{39.3M}} & 25.2s & 0.576  & 0.043 & 0.111 &  0.329 & 0.581 \\
			 
			\specialrule{0.75pt}{0.5pt}{0pt}
		\end{tabular}
	}
\end{table*}

Existing mainstream backbone networks can be divided into CNN-based and Transformer-based. In experiments, ResNet50, EfficientNet-B5, and ConvNext-T are adopted in the CNN-based category. Also, Transformer-based models are divided into two types: Unstructured and Hierarchical. The Unstructured denotes the type of native ViT architecture including ViT-S, ViT-B, DeiT-S, whereas Hierarchical refers to models possessing hierarchical architecture such as PvT-S and CvT13. To ensure fairness, all weights of backbones are pretrained from the timm~\cite{rw2019timm} framework, and the Post-Fusion structure with \textit{GC} is adopted.

The experimental results are presented in Table \ref{tab_backbone}, leading to several key conclusions. Firstly, Transformer-based backbones significantly outperform CNN-based ones. This superiority is largely attributed to the UAV self-localization task's inherent challenges, such as domain differences, rotational uncertainties, viewpoint offsets, and temporal variations in spatial information. These challenges necessitate networks that can globally model and extract contextual semantic information which is a strength of Transformer architectures. Secondly, even unstructured Transformer models exhibit relatively strong performance, likely due to the limited utilization of shallow features in this context. Among the hierarchical models, the CvT13 backbone network stands out, not only for its hierarchical feature output but also for its overall performance. Consequently, CvT13 is adopted as the backbone network for our baseline, as its hierarchical structure facilitates multi-scale feature fusion and demonstrates excellent performance.

\begin{table}[!t]
	\centering
	\tiny
	\renewcommand\arraystretch{1.0}
	\caption{The impact of different loss functions on performance.}
	\label{tab_loss}
	\resizebox{1.0\hsize}{!}{
		\begin{tabular}{c|c|ccc}
			\specialrule{0.75pt}{0pt}{0.5pt}
			Loss & RDS &  MA@5 &  MA@10 & MA@20 \\
			\specialrule{0.5pt}{0.5pt}{0.5pt}
			Focal Loss~\cite{bib31} & 0.730 & 0.261 & 0.578 & 0.794 \\
			CrossEntropy Loss~\cite{celoss}  &  0.750 & 0.276 & 0.606 & 0.828   \\
			Balance Loss~\cite{balanceloss} & 0.654 & 0.197 & 0.455 & 0.676  \\
			Weighted Balance Loss& \textbf{\underline{0.758}} & \textbf{\underline{0.298}} & \textbf{\underline{0.625}} & \textbf{\underline{0.837}} \\
			\specialrule{0.75pt}{0.5pt}{0pt}
		\end{tabular}
	}
\end{table}


\subsection{Loss Functions}\label{sec6.5}


\subsubsection{Loss Type}\label{sec6.5.1}
In the experiments, we trained with four different loss functions: cross-entropy loss, focal loss, balance loss, and the proposed weighted balance loss (WBL). The experimental results are shown in Table. \ref{tab_loss}. The performance of using balance loss alone is far worse than using cross-entropy loss or focal loss. As a result, a complete balance of positive and negative samples is not conducive to model convergence. Since the number of positive samples is much smaller than that of negative samples, the weight assigned to a single negative sample will be much smaller than that of positive samples after balancing. However, the results based on WBL lead to a 10-points RDS increase, and all MA@K indicators also achieve significant improvements.
\begin{figure}
	\centering
	\includegraphics[width=1.0\linewidth]{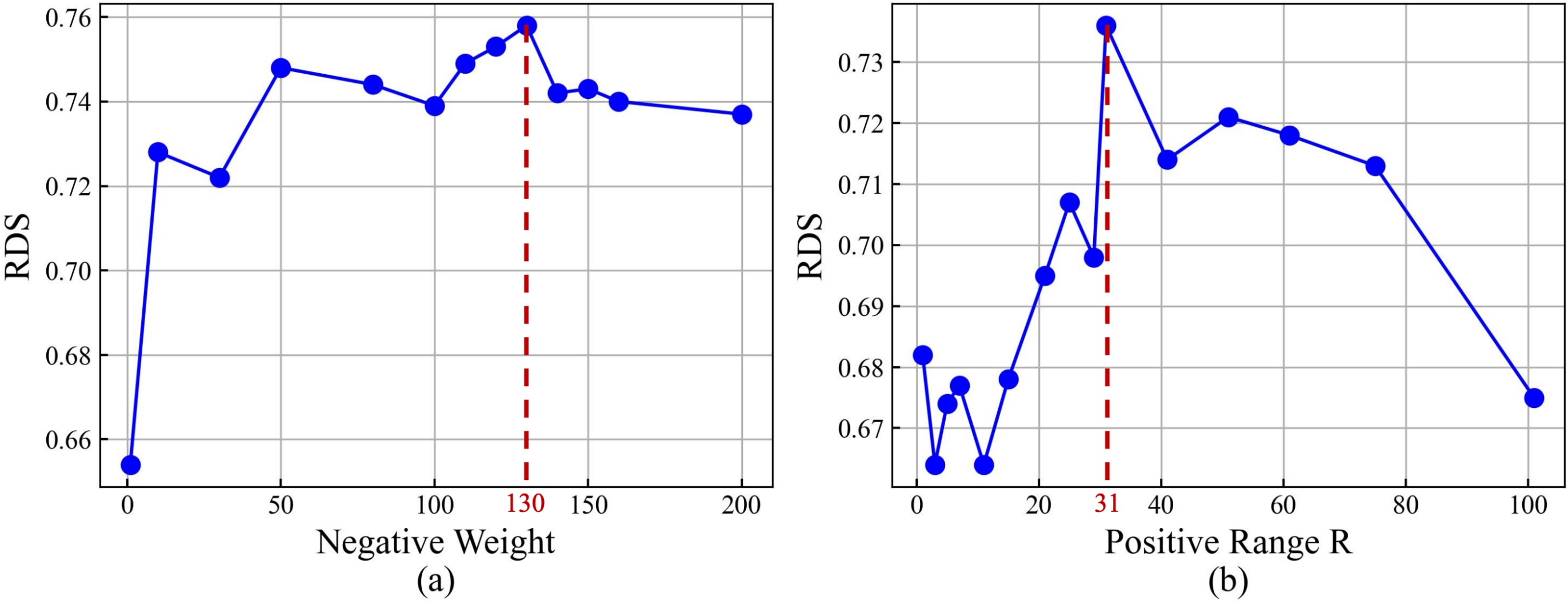}
	\caption{The impact of negative sample weight and positive sample sampling range on RDS index.}
	\label{figure_negW_posR}
\end{figure}
\begin{figure}
	\centering
	\includegraphics[width=1.0\linewidth]{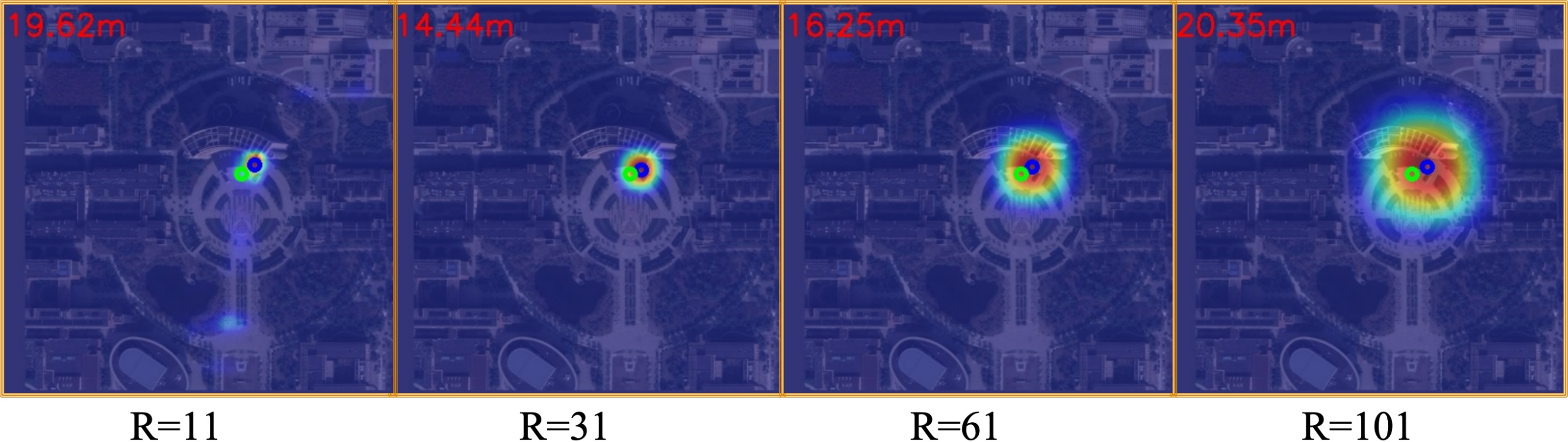}
	\caption{The influence of R on the thermal distribution of the final output heatmap. As R increases, the thermal distribution takes the form of diffusion.}
	\label{fig_heatmap_centerR}
\end{figure}

\subsubsection{Negative Weight}\label{sec6.5.2}
To find the most suitable value of $\textit{w}_\textit{neg}$, we made a lot of experiments with $\textit{w}_\textit{neg}$ as the variable, and the experimental results are shown in the Fig. \ref{figure_negW_posR}(a). When $\textit{w}_\textit{neg}=1$, the proposed WBL is equivalent to balance loss. When $\textit{w}_\textit{neg}$ reaches 130, all indicators are reaching the optimal state, as $\textit{w}_\textit{neg}$ keeps increasing, and all indexes gradually decrease. Last, $\textit{w}_\textit{neg}=130$ is adopted in our baseline.
\subsubsection{Positive Range}\label{sec6.5.2}
Positive range refers to the range occupied by positive samples in the feature map. Its setting method is shown in Fig. \ref{fig_centerR}. This experiment uses the $R$ parameter in both training and testing. In training, it represents the range of positive samples. In testing, the hanning window size is consistent with $R$. In fact, as shown in Fig. \ref{fig_heatmap_centerR}, the value of $R$ directly determines the degree of dispersion of the output heat map. The larger the $R$ value, the more divergent the heat distribution. From the visualization, it can be found that overly focused and overly divergent thermal distribution are unfavorable factors, so a moderate parameter $R$ needs to be selected to tune performance. Finally, we conducted experiments on different values of $R$, as shown in Fig. \ref{figure_negW_posR}(b). The optimal performance was achieved when $R=31$.

\subsection{Input Scale}\label{sec6.6}

\begin{figure}
	\centering
	\includegraphics[width=0.9\linewidth]{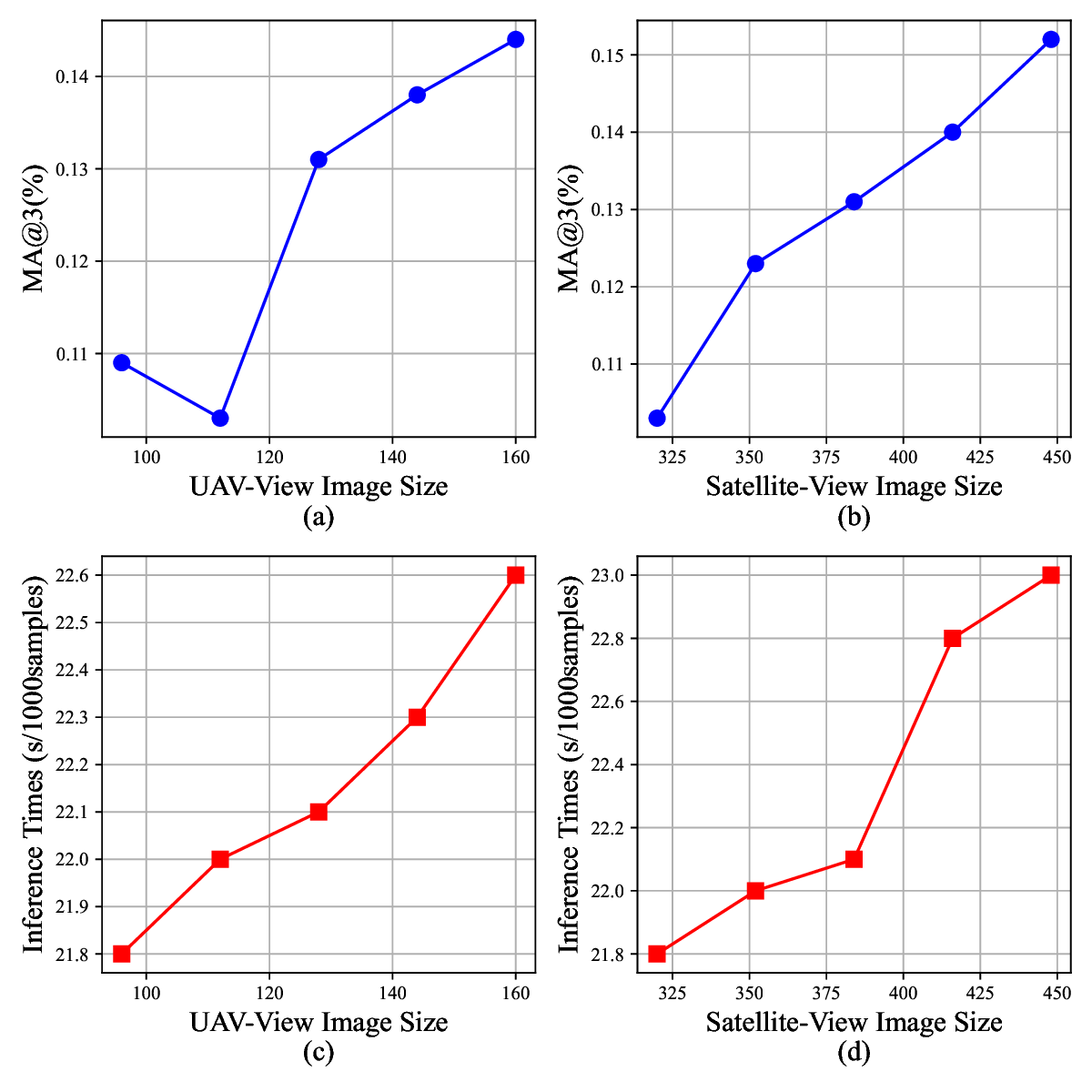}
        \vspace{-10pt}
	\caption{The impact of varied input scales on the performance.}
	\label{figure_input_scale}
\end{figure}

High-resolution input can provide more fine-grained information. We conduct two sets of comparison experiments on input images. One group is to control the UAV image scale as 128 pixels and change the satellite image scale (320$\rightarrow$448 pixels; stride=32), The other group is to control the satellite image scale as 384 pixels and change the UAV image scale (96$\rightarrow$160 pixels; stride=16). The results are shown in \ref{figure_input_scale}(a)(b), we can conclude that as the input scale increases, the positioning performance generally shows a positive correlation trend. Additionally, the inference time is counted in Fig. \ref{figure_input_scale}(c)(d). It can be seen that the effect of the image scale variation of UAV on the inference time is smaller, mainly because the image scale of UAV is 
much smaller compared to satellite. Last, the resolutions of satellite and UAV images used for our baseline are 384 and 128 pixels, respectively.

\subsection{Sharing Weight in Post-Fusion}\label{sec6.7}

\begin{table}[!t]
	\centering
	\tiny
	\renewcommand\arraystretch{1.0}
	\caption{Ablation experimental results of padding operation and shared weights. It can be observed that padding can play a positive role in both architectures.}
	\label{tab_share_padding}
	\resizebox{1.0\hsize}{!}{
		\begin{tabular}{c|cc|c|cc}
			\specialrule{0.75pt}{0pt}{0.5pt}
			
			Structure & Padding & Sharing & RDS & MA@3 & MA@20\\
			
			\specialrule{0.5pt}{0.5pt}{0.5pt}
			
			\multirow{3}{*}{Post-Fusion} & $\times$ & $\times$ & 0.407 & 0.016  & 0.226\\
			& \checkmark & $\times$ & \underline{\textbf{0.576}} & \underline{\textbf{0.111}}  & \underline{\textbf{0.581}}  \\
			& \checkmark & \checkmark & 0.276 & 0.003  & 0.093 \\
			
			\specialrule{0.5pt}{0.5pt}{0.5pt}
			
			\multirow{2}{*}{Mix-Fusion} & $\times$ & $\times$ & 0.571 & 0.049  & 0.493\\
			& \checkmark  & $\times$ & \underline{\textbf{0.680}}  & \underline{\textbf{0.137}} & \underline{\textbf{0.721}}  \\
			
			\specialrule{0.75pt}{0.5pt}{0pt}
		\end{tabular}
	}
\end{table}
Weight sharing can help reduce the parameters. As shown in Table \ref{tab_share_padding}, we find that the training process converges very slowly and performance is extremely poor when weights are shared. That is because the two-source data inputs have a massive domain difference, with uncertainties like viewpoint bias and rotation. Additionally, unlike the CVGL tasks, this task needs to fuse features and learn the spatial consistency relationship between two domains. Non-sharing allows for more differences in the learning process of the two domains.

\subsection{Padding in GC}\label{sec6.8}

Both structures of Post-Fusion and Mix-Fusion can use GC for feature fusion in their output stages. However, we find
that padding in GC plays a crucial role in maintaining spatial position information. The padding setting is to make the kernel center slide over each pixel of the source, ensuring that the spatial information of the output feature map is consistent with the original source. As shown in Table \ref{tab_share_padding}, the introduction of padding in GC has resulted in significant improvements (16.9-points boost in RDS for Post-Fusion and 10.9-points boost in RDS for Mix-Fusion).


\section{Visualization}\label{sec7}

\begin{figure}
    \centering
\includegraphics[width=0.9\linewidth]{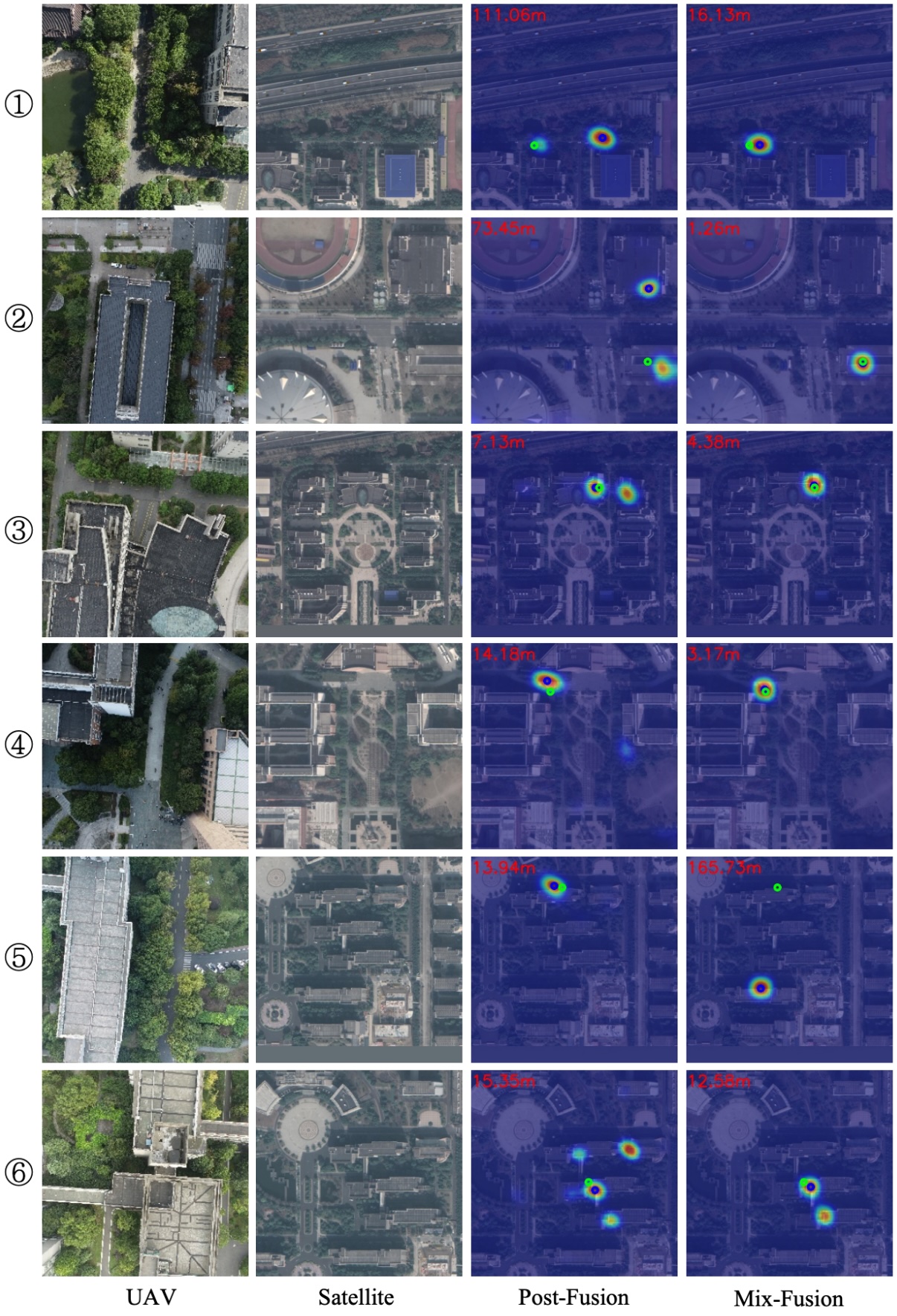}
 \caption{The first and second columns are the UAV and satellite images, and the third and fourth columns are the heatmaps of the Post-Fusion and Mix-Fusion, respectively. The green circle is groundtruth, and the blue is the predicted position. The red number in the upper left corner is the positioning error, and the unit is meter.}
	\label{figure_visualization}
\end{figure}

According to the characteristics of the output feature map supervision, the heatmap shows the spatial distribution of response strength after completing the sigmoid. That is, the higher the value in the heatmap, the greater the probability. Based on this setting, we normalize the output feature map and directly visualize it to analyze the response results. The visualization results are shown in Fig. \ref{figure_visualization}, which can be divided into 3 types of cases. From groups 1 and 2, it can be observed that the structure based on Mix-Fusion presents more positive results in thermal response, without ambiguous thermal distribution. We believe that this is mainly due to the deep feature interaction method of Mix-Fusion, which allows the model to better understand spatial semantic information and thus generate more confident results. From groups 3 and 4, it can be seen that the Mix-Fusion structure is better at small distance deviations because Mix-Fusion allows the backbone to learn relative spatial distribution information in shallow features, thereby more precise positioning. Last, groups 5 and 6 are cases where Mix-Fusion performed poorly. However, when we compared the original input UAV images, there was indeed a great deal of similarity in the locations of the mismatches. For such scenes with very similar images in space, it is indeed a huge challenge for the model to have stronger fine-grained mining and spatial information identification capabilities.

\section{Limitations and Future Work}\label{sec8}
There are still many aspects of DRL worthy of improvement. 1) Poor high-precision positioning: This is mainly limited by the large spatial distance between unit pixels of the output heat map. Adopting certain strategies to further expand the output heat map size, or learning an additional offset to further improve high-precision positioning performance is a direction worth exploring. 2) Overcoming the negative effects of scale and spatial distribution: As can be seen from Fig. \ref{figure_spatial_scale} and Fig. \ref{figure_query_distribution}(a)(b), changes in scale and spatial distribution currently have a negative impact on positioning, and design methods to mitigate it are also worth exploring.

\section{Conclusion}\label{sec9}

This paper introduces an efficient multi-source spatial feature interaction method, namely Drone Referring Localization (DRL), which not only interacts with heterogeneous features in a learnable manner and overcomes the inherent errors but also simplifies the entire localization process in an end-to-end manner. 
Meanwhile, two model structures named Post-Fusion and Mix-Fusion are proposed, which are implemented as dual-stream and single-stream networks, respectively. 
Experiments show that the deep fusion method (Mix-Fusion) can achieve better performance.
Furthermore, random scale crop (RSC) augmentation method is proposed to expand paired data with random scales and offsets. 
Also, weighted balanced loss (WBL) is designed to emphasize the influence of negative samples in the training process.
To train the DRL model, a paired dataset (UL14) containing UAV and satellite images is constructed. Additionally, two evaluation metrics are presented to assess positioning accuracy: meter-level accuracy (MA) and relative distance score (RDS). 
In addition, this paper carries out a large number of experimental analyses from the data and model levels to provide some references.
Finally, our proposed DRL architecture achieves better positioning performance (MA@20 +9.4\%) compared to IR-based methods, while exponentially reducing time (1/7) and storage (1/3) consumption.




\bibliographystyle{IEEEtran}
\bibliography{IEEEabrv,FPI}

\end{document}